\documentclass[10pt,twocolumn,letterpaper]{article}

\usepackage[accsupp]{axessibility}  % Improves PDF readability for those with disabilities.
\usepackage{iccv}
\usepackage{times}
\usepackage{epsfig}
\usepackage{graphicx}
\usepackage{amsmath}
\usepackage{amssymb}

\usepackage{graphicx}
\usepackage{amsmath}
\usepackage{amssymb}
\usepackage{booktabs}
\usepackage{stfloats}

\usepackage{overpic}
\usepackage[ruled,linesnumbered]{algorithm2e}

\usepackage{times}
\usepackage{epsfig}
\usepackage{graphicx}
\usepackage{amsmath}
\usepackage{mathrsfs}
\usepackage{amssymb}
\usepackage{multirow}
\usepackage{graphics}
\usepackage{threeparttable}
\usepackage{color}
\usepackage{framed}
\definecolor{shadecolor}{rgb}{.99,.91,.95}
\usepackage[normalem]{ulem}
\usepackage{float}
\usepackage{amsfonts}
\usepackage{bm}
\usepackage{bbm}
\usepackage{enumitem}
\usepackage[numbers]{natbib}
\usepackage{subcaption}
\usepackage{array}
\usepackage{colortbl}
\usepackage{pifont}
\usepackage[colorlinks]{hyperref}
\usepackage{mathtools}
\usepackage{siunitx}
\usepackage[nopar]{lipsum}
\usepackage{listings}
\usepackage[export]{adjustbox}
\usepackage{verbatim}
\usepackage{colortbl}

\usepackage{bbding}
\usepackage{pifont}
\usepackage{wasysym}
\usepackage{amssymb}

\usepackage[ruled,linesnumbered]{algorithm2e}
\usepackage[dvipsnames]{xcolor}
\newcommand{\stdvu}[1]{\scriptsize{\color{darkgray}(#1)} {\color{ForestGreen}$\uparrow$}}

\definecolor{mypink}{rgb}{.99,.91,.95}
\usepackage{etoolbox}
\makeatletter
\usepackage[dvipsnames]{xcolor}

\newcommand{\stdvd}[1]{\scriptsize{\color{darkgray}(#1)} {\color{red}$\downarrow$}}

\newcommand{\stdvno}[1]{\scriptsize{\color{darkgray}(#1)} {\color{mygray}$\downarrow$}}
\definecolor{firebrick}{rgb}{0.7, 0.13, 0.13}
\definecolor{darkpastelgreen}{rgb}{0.01, 0.75, 0.24}
\definecolor{deepskyblue}{rgb}{0.0, 0.75, 1.0}
\definecolor{mypink2}{rgb}{.99,.96,.98}%浅
\definecolor{mypink1}{rgb}{.99,.93,.98}
\definecolor{mypink}{rgb}{.99,.90,.98}%深
\definecolor{mygray}{rgb}{.95,.95,.95}
\definecolor{lv14}{rgb}{0.5,0.5,0.5}
\let\@algcomment\relax

\newcommand\algcomment[1]{\def\@algcomment{\footnotesize#1}}
\renewcommand\fs@ruled{\def\@fs@cfont{\bfseries}\let\@fs@capt\floatc@ruled
	\def\@fs@pre{\hrule height.8pt depth0pt \kern2pt}%
	\def\@fs@post{}%
	\def\@fs@mid{\kern2pt\hrule\kern2pt}%
	\let\@fs@iftopcapt\iftrue}
\makeatother
\usepackage{tabulary}
\newcolumntype{I}{!{\vrule width 1pt}}

\newcolumntype{x}[1]{>{\centering\arraybackslash}p{#1pt}}
\newcolumntype{y}[1]{>{\raggedright\arraybackslash}p{#1pt}}
\newcolumntype{z}[1]{>{\raggedleft\arraybackslash}p{#1pt}}
\newlength\savewidth

\iccvfinalcopy % *** Uncomment this line for the final submission

\def\iccvPaperID{1685} % *** Enter the ICCV Paper ID here

\ificcvfinal\fi

\begin{document}

%%%%%%%%% TITLE
\title{Diverse Data Augmentation with Diffusions for Effective \\Test-time Prompt Tuning}
\author{Chun-Mei Feng$^1$\quad Kai Yu$^1$\thanks{Corresponding author.}\quad Yong Liu$^1$\quad Salman Khan$^{2,3}$ \quad Wangmeng Zuo$^4$ \\
$^1$Institute of High Performance Computing (IHPC), \\Agency for Science, Technology and Research (A*STAR), Singapore\\
$^2$Mohamed bin Zayed University of Artificial Intelligence (MBZUAI), UAE\\
$^3$Australian National University, Canberra ACT, Australia\\
$^4$Harbin Institute of Technology, Harbin, China\\
{\tt\small fengcm.ai@gmail.com; yu\underline{ }kai@ihpc.a-star.edu.sg}
\\ \small {\url{https://github.com/chunmeifeng/DiffTPT}}
}
% \and

\maketitle
\ificcvfinal\thispagestyle{empty}\fi

%%%%%%%%% ABSTRACT
\begin{abstract}

Benefiting from prompt tuning, recent years have witnessed the promising performance of pre-trained vision-language models, e.g., CLIP, on versatile downstream tasks.
% In this paper, we consider a particular setting of learning adaptive prompt on the fly for each test sample from unseen new domain, i.e., test-time prompt tuning (TPT). 
In this paper, we focus on a particular setting of learning adaptive prompts on the fly for each test sample from an unseen new domain, which is known as test-time prompt tuning (TPT).
%
%Nonetheless, it remains difficult for the model to adapt to unseen data during testing, especially the natural distributional variation that occurs in real-world deployment scenarios. 
%
% To this end, existing TPT method usually resort to data augmentation and confidence selection.
Existing TPT methods typically rely on data augmentation and confidence selection.
However, conventional data augmentation techniques, e.g., random resized crops, suffers from the lack of data diversity, while entropy-based confidence selection alone is not sufficient to guarantee prediction fidelity. 
%
%Standard data augmentation methods, \eg, rotation and flipping, are still suboptimal in the face of natural distribution shifts due to a lack of diversity along specific semantics. 
%
% To enforce data diversity and consistency, we present a novel TPT method, dubbed DiffTPT, by leveraging pre-trained diffusion model to generate diverse and informative new data.
To address these issues, we propose a novel TPT method, named DiffTPT, which leverages pre-trained diffusion models to generate diverse and informative augmented data.
%
%To fill this gap, this paper proposes a method for augmenting a single test sample with class-consistent semantics and increased content diversity, as well as optimizing text prompts, dubbed DATPT, to cope with natural distributional variations in the real world. 
%
Specifically, we incorporate augmented data by both conventional method and pre-trained stable diffusion to exploit their respective merits, improving the model's ability to adapt to unknown new test data.
% making the model better adapt to unknown new test data. 
%
%DATPT is based on the major insight that creating diverse new samples by optimizing the latent features of the test samples more accurately represents the natural distribution drift, making the model adapt to unknown data during testing. 
%
Moreover, to ensure the prediction fidelity of generated data, we introduce a cosine similarity-based filtration technique to select the generated data with higher similarity to the single test sample.
% Furthermore, to encourage the prediction fidelity of generated data, {\color{red}we introduce a cosine similarity based  by selecting the generated data with higher similarity with the single test sample.}
%
%We first create new samples with rich information through pre-trained stable diffusion and then optimize the prompt by maximizing cosine similarity with the single test sample, thereby keeping the class semantics of the generated samples unchanged. 
%
Our experiments on test datasets with distribution shifts and unseen categories demonstrate that DiffTPT improves the zero-shot accuracy by an average of 5.13\% compared to the state-of-the-art TPT method.
% Experiments on the test datasets with distribution shifts and unseen categories show that, DiffTPT improves the zero-shot accuracy by an average of $5.13\%$ in comparison to the state-of-the-art test-time prompt-tuning method, TPT. 
% \textit{Our code and models will be publicly released.}

\end{abstract}

\vspace{-3pt}
\section{Introduction}\label{sec:intro}

Pre-trained vision-language models, such as CLIP~\cite{radford2021learning}, have recently demonstrated promising performance on a variety of downstream tasks without the need for task-specific training data~\cite{zhou2022learning,zhou2022conditional,li2022grounded,ramesh2022hierarchical}. 

\begin{figure}[t]
	% \vspace{-14pt}
	\begin{center}
		\includegraphics[width=\linewidth]{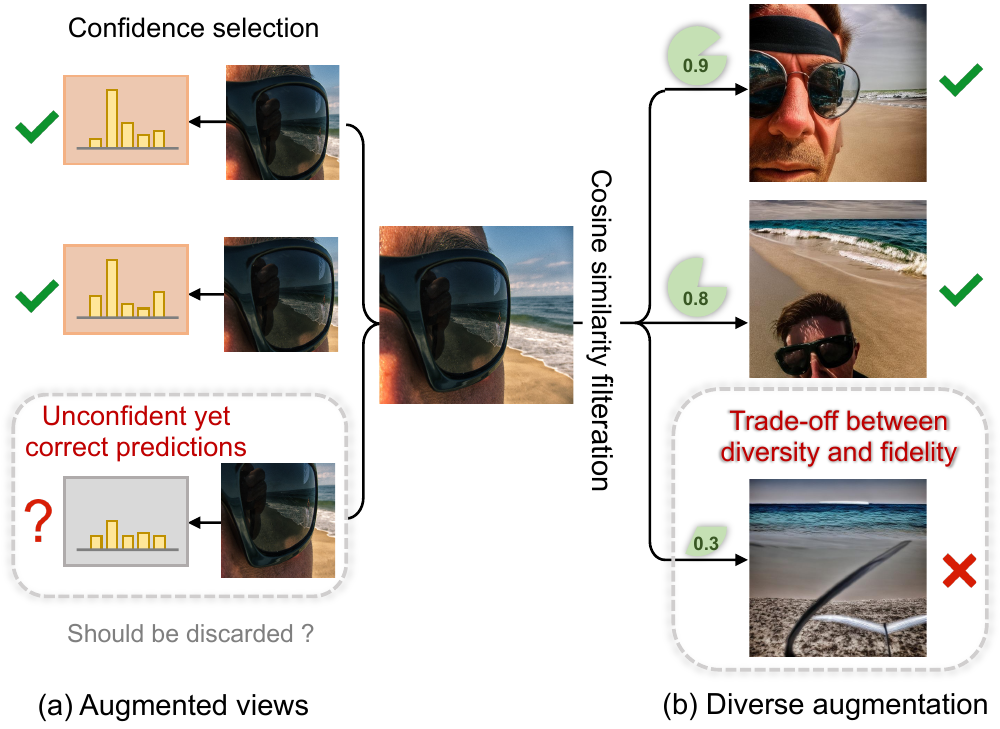}
	\end{center}
	\vspace{-15pt}
	\captionsetup{font=small}
	\caption{\small\textbf{(a)} Prior TPT method~\cite{shu2022test} uses different augmented views along with confidence selection, resulting in \textit{overly simplistic variants} in the test data and \textit{unconfident yet correct} predictions being discarded. In comparison, \textbf{(b)} our DiffTPT is effective in generating data with \textit{\textbf{richer visual appearance variation}} and selecting generated data with higher \textbf{\textit{prediction fidelity}}.}
	\vspace{-15pt}
	\label{fig:1}
\end{figure}
This is achieved through well-designed prompts, but both hand-      crafted prompts and prompt tuning have limitations as they are restricted to the training data distribution of the current domain, making it challenging to generalize to distributions outside the domain, especially in the zero-shot setting~\cite{mandal2019out}.
% This is mainly achieved by properly designed prompts. 
% However, neither hand-crafted prompts nor prompt tuning seem to be sub-optimal because they are all limited to the training data distribution of the current domain, making it difficult to generalize to distributions outside the domain, especially in the zero-shot setting~\cite{mandal2019out}.
% One is hand-crafted prompts, which require domain-specific heuristics and thus need to be carefully designed~\cite{}. The other one is prompt tuning; while it produces better results than hand-crafted, it is limited to the distribution of training data for the current task, making it difficult to generalize to other distributions, such as the zero-shot task~\cite{}. 
% Latest work has turned to test-time prompt tuning (TPT), which dynamically tunes prompts using only test examples without any training data or annotations, showing advantages in zero-shot generalization~\cite{shu2022test}. 
% In this work, we concentrate on TPT, which is more applicable for real-world scenarios since it is often challenge to collect a significant amount of labeled data for each different down-stream task.
In this context, test-time prompt tuning (TPT)~\cite{shu2022test} has been proposed to learn adaptive prompts on the fly for each test sample from an unseen new domain, without any training data or annotations. This setting offers a more practical approach for dynamic real-world scenarios where collecting a significant amount of labeled data for an unseen target distribution is often challenging.
% Following~\cite{shu2022test}, we consider a particular setting of learning adaptive prompts on the fly for each test sample from unseen new domain while without any training data or annotations, \ie, test-time prompt tuning (TPT). 
% Such setting is more practical for dynamic real-world scenarios since it is often challenging to collect a significant amount of labeled data for unseen target distribution. 
% downstream task.

% heavily basic invariants

One initial attempt to tackle TPT is to incorporate confidence selection with entropy minimization for prompt tuning by using various augmented views of each test sample~\cite{shu2022test}. 
%
%and minimizing the confidence choice entropy to make the model have consistent predictions~\cite{shu2022test}. 
%
However, the data augmentation method employed in \cite{shu2022test} uses simple parametric transformations to alleviate the scarcity of data (see Fig.~\ref{fig:1} (a)). Such simplistic transformations are limited in generating diverse data to reflect the rich appearance variation of the test sample~\cite{antoniou2017data,perez2017effectiveness,zhao2020differentiable,shorten2019survey}. 
%
%which lead to changes in the surface visual characteristics of the image but are invariant to the actual image content (see Fig.~\ref{fig:1}(a))~\cite{antoniou2017data,perez2017effectiveness,zhao2020differentiable}. 
%
% And the insufficient diversity in augmented data may weaken the generalization ability of the learned prompt.  
Insufficient diversity in the augmented data can hamper the generalization ability of the learned prompt leading it to overfit on a single data mode.
Also, entropy-based confidence selection proposed in \cite{shu2022test} is not enough to ensure prediction fidelity, as an augmented sample with low-entropy prediction could still be misclassified into a different class to the test sample, leading to unrepresentative samples in the augmented pool.
% As for confidence selection, Shu \etal~\cite{shu2022test} suggested to filter out the augmented samples with high-entropy predictions. 
% %
% However, entropy-based confidence selection is not sufficient to guarantee prediction fidelity. 
% For example, an augmented sample with low-entropy prediction will be kept even it is classified into a different class to that of the test sample.  

%The new information brought by such a method is limited because the standard augmented transformation choices respect the basic invariants that are present in the data~\cite{shorten2019survey}. As a result, the direct damage to zero-shot generalization is that it makes it vulnerable to \textit{natural distributional variations} in the real world.

% Recent progress in image generation has make it feasible to better handle the diversity of augmented data.
Recent advancements in image generation have made it possible to handle the diversity of augmented data better.
%
%Training models using synthetic images are currently evolving rapidly, as it offer greater flexibility than standard data augmentation methods. 
%
Early image generation methods, including VAEs~\cite{kingma2013auto} and GANs~\cite{goodfellow2020generative}, often require a large amount of data for training. 
Recently, diffusion models have achieved superior performance in text-to-image generation with high-quality photo-realistic details~\cite{nichol2021glide,ramesh2022hierarchical,saharia2022photorealistic,rombach2022high}.  
In comparison to the data augmentation method adopted in~\cite{shu2022test}, the augmented data by diffusion model can exhibit much higher diversity, thereby providing richer
visual representations and benefiting  the generalization ability of
learned prompts.  
However, diffusion-based data augmentation is prone to producing \textit{spurious} augmentations which are more difficult to be filtered out by entropy-based confidence selection.
Thus, further research is required to find the right \textit{balance} between data diversity and prediction fidelity when applying diffusion-based data augmentation for TPT.
% Thus, further studies on the \textit{trade-off between data diversity and prediction fidelity} is required to apply  diffusion-based data augmentation for TPT. 

%the advantage of generating high-quality images with more realistic details, which motivates us to directly augment the test data with diffusion models, making the test-time optimization more robust to natural distribution variations~\cite{ruiz2022dreambooth,preechakul2022diffusion,gu2022vector}. Compared to~\cite{shu2022test}, this approach is not limited to generating a fixed, limited set of hand-designed augmentations and can modify image appearance in a way that respects object-level visual invariance, providing richer visual representations for test samples while preserving semantic categories that are unchanged (see Fig.~\ref{fig:1}(b)). While seemingly straightforward, training a model with images generated by a diffusion model requires a \textit{trade-off between diversity and fidelity}; otherwise, an overemphasis on synthetic data can introduce spurious augmentations that collapse the model~\cite{nichol2021glide}. For example, Shu \textit{et al.} optimized the prompt by applying entropy-based confidence selection, which cannot fully guarantee data fidelity as it not only discards \textit{unconfident yet correct predictions} but also does nothing for images that are differ from the original samples, \eg, \textit{spurious augmentations}~\cite{shu2022test}.

In this work, we introduce a new TPT method called DiffTPT, that enhances the \textit{\textbf{data diversity}} of test samples through diffusion models while maintaining \textit{\textbf{prediction fidelity}} with cosine similarity-based filtration (see Fig.~\ref{fig:1} (b)).
%
%to cope with \textit{\textbf{natural distribution variations}} in the real world. 
%
In terms of data diversity, DiffTPT adopts Stable Diffusion for data augmentation. 
Stable Diffusion is a text-to-image generation model which synthesizes an image based on the CLIP text feature~\cite{rombach2022high}.
Instead, we use the CLIP image feature of the test sample as an alternative to the CLIP text feature and feed it into Stable Diffusion for data augmentation.
The diffusion-based augmentation is effective in generating diverse images with \textit{richer visual appearance variation} while \textit{\textbf{preserving the key semantics}}.
% Benefited from the generation ability of Stable Diffusion, diffusion-based augmentation is effective in generating diverse images with richer visual appearance variation while faithfully preserving the key semantics.
%
%DiffTPT is based on the insight that by optimizing the latent features of the test samples to create new samples with diverse images, making the model accurately characterize the natural distribution drift and adapt to unknown data during testing. 
%
Furthermore, we leverage both the augmentation in~\cite{shu2022test} and diffusion-based augmentation for improving TPT performance. 
To ensure prediction fidelity, we introduce cosine similarity-based filtration to \textit{remove spurious augmentations}. %
By incorporating diffusion-based augmentation with cosine similarity-based filtration, our DiffTPT can make a fair {{trade-off between diversity and fidelity}}. 
Moreover, DiffTPT is agnostic to the training data and can be seamlessly integrated into arbitrary CLIP architectures.
%
%In particular, we absorb the merits of both standard augmentations and pre-trained stable diffusion, leveraging the model robust to unknown new test data. 
%
%To make a fair \textit{\textbf{trade-off between diversity and fidelity}} of the generating samples, DiffTPT optimizes the prompt by \textit{cosine similarity based } with the single test sample, to remove spurious augmentations that may be introduced by \textit{\textbf{overemphasizing} synthetic data}. DiffTPT is a simple but effective method that is agnostic to the training data and can be seamlessly integrated into arbitrary CLIP architectures. 
Experimental results show that DiffTPT achieves a notable improvement of zero-shot accuracy by an average of $5.13\%$ in comparison to the state-of-the-art TPT method~\cite{shu2022test}. 
To sum up, our contributions are as follows:
\begin{itemize}[leftmargin=*]
	\setlength{\itemsep}{0pt}
	\setlength{\parsep}{-2pt}
	\setlength{\parskip}{-0pt}
	\setlength{\leftmargin}{-15pt}
	\vspace{-6pt}
\item 
   We present a new test-time prompt tuning method, \ie, DiffTPT, based on a pre-trained diffusion model, that balances the trade-off between data diversity and prediction fidelity.
\item 
   Diffusion-based data augmentation is proposed to generate diverse augmented images with richer visual appearance variations while faithfully preserving the key semantics.
   %
   %{We propose a new TPT algorithm that towards diverse data augmentation, DiffTPT, to improve zero-shot generalization under \textit{natural distributional variations}. By using a pre-trained diffusion model to modify test image appearance in a way that respects \textit{object-level visual invariance}, DiffTPT presents the key semantics of the test samples in a variety of visual representations for test samples.}
\item 
   We introduce cosine similarity filtration to remove {spurious augmentations}, thereby improving the prediction fidelity of augmented images.
   %
   %{We explore the \textit{trade-off} between diversity and fidelity of newly generated test samples by a cosine similarity  with the single test sample, making DiffTPT to remove \textit{spurious augmentations}.}
   % that may be introduced by overemphasizing synthetic data
\item 
   Experimental results show that our DiffTPT significantly outperforms the state-of-the-art test-time prompt-tuning method~\cite{shu2022test}.
   %
   %{We evaluate the performance of DiffTPT on zero-shot generalization via 
   %natural distribution shifts and cross-dataset generalization
   %$various experiments, results show that DiffTPT significantly outperform the state-of-the-art test-time prompt-tuning %method~\cite{shu2022test}. Ablations confirm that our method achieves a fair trade-off between diversity and fidelity of the generated test image.   }
   
\vspace*{-7pt}
\end{itemize}

% \vspace{-3pt}
\section{Related Work}
\noindent\textbf{Prompt Tuning.} 
Large-scale pre-trained models have improved performance on several tasks in natural language processing~\cite{devlin2018bert,radford2018improving} and computer vision~\cite{jia2021scaling,chen2020simple,jia2022visual,feng2023learning,li2022prompt} by learning general representations and transferring the learned knowledge to downstream tasks. 
For adapting pre-trained models to downstream tasks, several parameter-efficient fine-tuning methods, \eg, prompt tuning and adapters, have been proposed in the recent few years.
%
% Among them, prompt tuning enables pre-trained models to directly adapt to downstream tasks by adding a few trainable tokens to the input. 
%
For example, CoOp~\cite{zhou2022learning} and CoCoOp~\cite{zhou2022conditional} employ continuous prompt optimization strategies and instance-wise conditionalization on prompts to achieve generalization to out-of-distribution data. CLIP-Adapter~\cite{gao2021clip} and Tip-Adapter~\cite{zhang2021tip} use adapters and non-parametric key-value cache models to fine-tune the CLIP model, thereby improving its adaptability to the target dataset.
%
% For example, CoOp adopted a continuous prompt optimization strategy to avoid labor-intensive prompt engineering~\cite{zhou2022learning}. Further, CoCoOp conditioned prompt in an instance-wise manner to alleviate the lack of generalization ability of CoOp for out-of-distribution data~\cite{zhou2022conditional}. 
% CLIP-Adapter introduced a feature adapter to fine-tune CLIP model, making the model learn new features while maintaining a simple design~\cite{gao2021clip}. 
% Tip-Adapter employed a non-parametric key-value cache model to train the adapter, avoiding backpropagation and improving the adaptability to the target dataset~\cite{zhang2021tip}. 
%
UPL alleviated CLIP's reliance on labeled data and trains a prompt representation ensemble to improve transfer performance without the label of target dataset~\cite{huang2022unsupervised}. However, the performance of zero-shot generalization is highly dependent on well-designed prompts. %, \eg, an appropriate set of prompts, which would otherwise decrease accuracy. 

In another line of work,~\cite{shu2022test} proposed test-time prompt tuning (TPT) by generating multiple random augmented views of a single test sample that is directly applicable to the zero-shot generalization of the base model~\cite{shu2022test}. However, the data augmentation in~\cite{shu2022test} suffers from overly simplistic variants and the entropy-based confidence selection is not sufficient to guarantee prediction fidelity.
%
%this approach \textit{strictly} follows the basic invariants that are present in the test data, making zero-shot generalizations vulnerable to \textit{natural distributional variations} in the real world. 
%
To improve TPT~\cite{shu2022test}, our work suggests incorporating diffusion-based data augmentation and  cosine similarity-based filtration for a better trade-off between data diversity and prediction fidelity.

%diverse new samples to characterize the natural distribution drift by optimizing the latent features of the test samples, making the model adapt to unknown data during testing.

% \vspace{-2pt}
\noindent\textbf{Test-time Optimization.}~Adapting machine learning models to test samples is a more challenging and practical setting where no training data is available during inference~\cite{wang2020tent,sun2020test,chen2022contrastive,shanmugam2021better}. This setting alleviates the limitation of inaccessible source data due to privacy concerns and enables training the model once and adapting it to any unknown test distributions~\cite{gao2022visual}. 
To design an efficient test-time objective, one way is to make the objective independent of a specific training procedure by minimizing the entropy of the batch prediction probability distribution~\cite{wang2020tent} or bypassing the requirement of multi-test samples via data augmentation~\cite{zhang2021memo}. 
%
% For example, an extra branch is introduced to adapt the network to test samples by optimizing the objective at test time~\cite{sun2020test,liu2021ttt++}. 
% By allowing the model to adapt itself using feedback from its own predictions, Wang \textit{et al.}~\cite{wang2020tent} optimized the confidence of the model at test time by minimizing generalization error on shifted data. 
%
% However, the number of test samples required for the model to output a non-trivial solution limits this approach. 
%
Another approach is to explicitly apply the BN layer at test time to constrain a set of parameters for optimization and enhance the robustness of the model to distribution shifts ~\cite{schneider2020improving}. 
%
% However, such mechanisms limit the scalability of the model architecture. 
%

However, these methods are either limited by the number of test samples required for the model to output non-trivial solutions or by the scalability of the model architecture.
Subsequent works moved to large-scale, pre-trained models with parameter-efficient tuning~\cite{gao2022visual,zhang2022tempera}. For example, TPT learned target-specific text prompts while freezing the backbone by generating multiple randomly augmented views during the test phase and filtering out noise augmentations that may lead to misleading predictions through entropy minimization~\cite{shu2022test}. 
However, the entropy-based confidence selection~\cite{shu2022test} is limited in its ability to filter out a misclassified augmented sample with low entropy prediction. 
%
%we find that \cite{shu2022test} discards a large number of \textit{unconfident yet correct predictions} (see Fig.\ref{fig:1} (a)). 
%
Given this, we introduce a cosine similarity-based filtration between augmented and test samples (see Fig.\ref{fig:1} (b)) to encourage the augmented samples to preserve consistent class semantics (\ie, \textit{prediction fidelity}) while bringing more \textit{diverse} information.

% \vspace{-5pt}
\noindent\textbf{Image Synthesis.} 
Training models with synthetic images is gaining popularity and undergoing rapid development. 
In contrast to standard data augmentation methods, such as image manipulation~\cite{shorten2019survey}, image erasing~\cite{zhong2020random}, and image mixup~\cite{zhang2020does,hendrycks2019augmix}, image synthesis offer higher flexibility as these methods augment datasets with pre-defined transformations and cannot provide images with highly diverse content. Early image generation methods, including VAEs~\cite{kingma2013auto} and GANs~\cite{goodfellow2020generative}, initially provided promising generated images~\cite{brock2018large}, and have been widely applied to various vision tasks. 
Most recently, diffusion models have been developed to generate higher-quality images with more photo-realistic details than the prior image generation methods~\cite{ho2020denoising,nichol2021improved,saharia2022photorealistic,ramesh2022hierarchical,zhang2022motiondiffuse}. 
Recent works have shown the outstanding performance of diffusion generative models in many applications, \eg, using the latent space of powerful pretrained autoencoders for high-resolution image synthesis~\cite{rombach2022high}, enhancing text-conditional image synthesis~\cite{nichol2021glide,ramesh2022hierarchical}, learning diffusion-based prior for few-shot conditional image generation~\cite{sinha2021d2c}, and probabilistic model for point cloud generation~\cite{ho2022imagen}.
% Recent works have shown the outstanding performance of diffusion generative models in many applications. 
% %
% For example, by incorporating cross-attention layers into the model, Rombach \textit{et al}. converted diffusion models into adaptable generators that can create high-resolution images from the latent space of powerful, pre-trained autoencoders~\cite{rombach2022high}. Nichol \textit{et al}.~\cite{nichol2021glide} and Ramesh \textit{et al}.~\cite{ramesh2022hierarchical} leveraged these representations from diffusion models to enhance text-conditional image synthesis. 
% %
% Similarly, diffusion-based prior can also be learned for few-shot conditional image generation~\cite{sinha2021d2c} and point cloud generation~\cite{ho2022imagen}. 
%Accordingly, the diffusion model has been proved by extensive work that can generate higher-quality images. 
These works motivate us to directly augment the test data with the \textit{same semantics but diverse information} through the diffusion model, thereby improving test-time prompt-tuning performance.

\begin{figure*}[t]
    % \vspace{-8pt}
	\begin{center}
		\includegraphics[width=\linewidth]{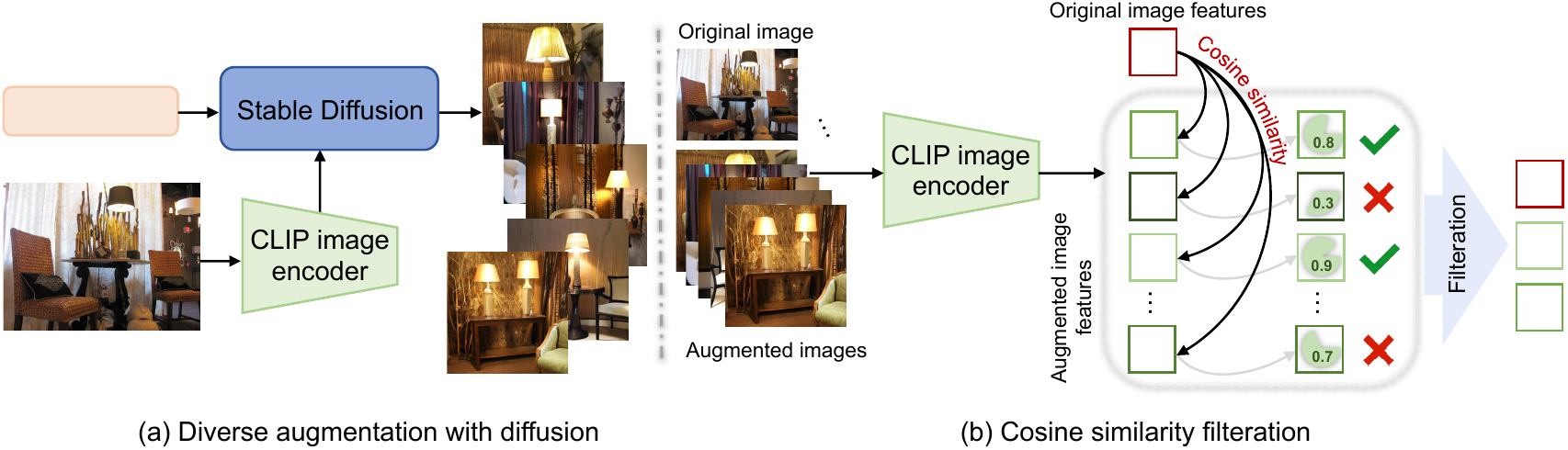}
        \put(-494,105){ \small$\mathbf{n} \sim \mathcal{N}(0, I)$}
        % \put(-108,62){ \small$\rho_{C}$}
	\end{center}
	\vspace{-15pt}
	\captionsetup{font=small}
	\caption{\small{\textbf{Overview} of our proposed \textbf{DiffTPT}. We first \textbf{(a)} use the pre-trained stable diffusion to generate data with \textit{richer visual appearance variation}, then \textbf{(b)} uses a cosine similarity based filtration with the single test sample to \textit{remove spurious augmentations}, making our method a \textbf{\textit{trade-off between diversity and fidelity}}.}}
	\vspace{-17pt}
	\label{fig:2}
\end{figure*}

% \vspace{-3pt}
\section{Methodology}\label{sec:method}
% \vspace{-2pt}
\subsection{Test-time Prompt Tuning}\label{sec:tpt}
The pre-trained vision-language models, \eg, CLIP, consist of two encoders, \ie, image encoder $f(\cdot)$ and the text encoder $g(\cdot)$, providing rich knowledge for various downstream tasks. For zero-shot classification, we can obtain the predication probability by 
\vspace{-4pt}
\begin{equation}
{p}(y_i \mid \mathbf{x})=\frac{\exp \left(\cos \left(\boldsymbol{w}_{\boldsymbol{i}}, \boldsymbol{e}\right) / \tau\right)}{\sum_{j=1}^K \exp \left(\cos \left(\boldsymbol{w}_{\boldsymbol{j}}, \boldsymbol{e}\right) / \tau\right)},
\end{equation}
where $\boldsymbol{e}$ is the image features extracted by $f(\cdot)$ for the image $\mathbf{x}$ that together with their paired text feature $\boldsymbol{w}_{{i}}$ are used to compute cosine similarity $\cos (\boldsymbol{w}_{{i}}, \boldsymbol{e})$ for class $i$. And $\tau$ is the temperature parameter.
%\in K

However, the performance of CLIP towards zero-shot generalization needs to be improved because the foundation model is often desired to generalize to out-of-distribution samples. 
Here, we consider test-time prompt tuning since it can modify the context of class names to adapt to new test-time data samples. Specifically, this means that only a test sample is available, and no other labeled training data is available for adaptation. Therefore, we need to optimize prompts based on a single test sample $\mathbf{x}_\texttt{test}\in \mathbb{R}^{C \times H \times W}$ during the testing phase. Formally, we have
\vspace{-3pt}
\begin{equation}
{\boldsymbol{v}^*}=\arg \min _{\boldsymbol{v}} \mathcal{L}\left(\mathcal{F}, \boldsymbol{v}, \mathbf{x}_{\texttt{test}}\right),
\vspace{-5pt}
\end{equation}
where $\mathcal{F}$ is the CLIP model consist of an image encoder $f(\cdot)$ and a text encoder $g(\cdot)$, $\boldsymbol{v}^*$ denotes the learnable prompts
of each class name $y_i$, which together form the category-specific text inputs $\left\{\boldsymbol{v}^* ; y_i\right\}$. For the classification task, $\mathcal{L}$ indicates the cross-entropy loss.

To enforce the effectiveness of TPT, Shu \textit{et al}.~\cite{shu2022test} use  multiple (\ie, $N$) augmented views along with the confidence selection mechanism
% and minimizing the entropy of the averaged prediction probability
, which can be expressed as: 
\vspace{-6pt}
\begin{equation}
\begin{aligned}
{\boldsymbol{v}^*}\!=\arg\min _{\boldsymbol{v}}-\sum_{i=1}^K \tilde{p}_{\boldsymbol{v}}\left(y_i \!\mid \!\mathbf{x}_{\texttt {test\!}}\right) \log \tilde{p}_{\boldsymbol{v}}\left(y_i \!\mid \!\mathbf{x}_{\texttt {test}}\!\right), 
\end{aligned}
\label{tpt0}
\end{equation}
\vspace{-7pt}
\begin{equation}
\tilde{\boldsymbol{p}}_{\boldsymbol{v}}=\frac{1}{\rho_H N} \sum_{n=1}^N \!\mathbbm{1}\!\left[\mathbf{H}\left(\boldsymbol{p}_n\right) \leq \tau\right] \boldsymbol{p}_{\boldsymbol{n}}\left(y \!\mid \!\mathcal{A}_n\left(\mathbf{x}_{\texttt {test}}\right)\right), 
\label{tpt}
\end{equation}
where $K$ denotes the number of classes. ${p}_{\boldsymbol{p}}\left(y_i \mid \mathcal{A}_i\left(\mathbf{x}_{\texttt {test}}\right)\right)$ represents the class probabilities for the $n$-th augmented view and prompt $\boldsymbol{v}$, $\tau$ is the confidence selection threshold resulting in a $\rho_H$ percentage on the total $N$ augmented views, $\mathbf{H}$ calculates the self-entropy of the prediction on an augmented view.

\subsection{Approach Overview}\label{sec:moti}
While the augmentation method in~\cite{shu2022test} has achieved considerable success in TPT, it is obvious that the solution depends heavily on the diversity of augmented images. Since the augmented views generally have the same object and background appearance content, the model suffers from overly simplistic variants in the test data, which can cause prompt overfitting.
%the basic invariants present in the augmented data. 
%
Moreover, Shu \textit{et al}.~\cite{shu2022test} use an entropy-based confidence selection mechanism to discard augmented images with high entropy prediction. That is, most of the retained augmented images are {cropped variants} of the objects in the original test image (see the augmented views of Fig.~\ref{fig:1} (a)). 
As a result, the augmentation method in~\cite{shu2022test} will result in {trivial variations in augmented images}, thereby limiting the generalization ability of the learned text prompt~\cite{bansal2023leaving}.

In this work, we circumvent this problem by leveraging a diffusion model on each test sample to generate diverse new images that {{capture natural variations in appearance}} while preserving the key semantics. 
%
%This approach generates new samples in a way that respects object-level variations and presents the {key semantics} of the test samples in a diverse {variety of visual representations} in the newly generated data. 
%
Thus, diffusion-based data augmentation not only increases the number of original test samples but also achieves semantic consistency in distribution variations. 
While simply applying diffusion to test-time prompt tuning allows for performance improvements (see Table~\ref{tab:1} in Sec.~\ref{sec:acc}), it may introduce spurious augmentations and can lead to wrong predictions. Furthermore, we address this issue by introducing a cosine similarity-based filteration. 
In the following, we will introduce diffusion-based data augmentation and cosine similarity-based filteration in more detail.

\subsection{Diffusion-based Diverse Data Augmentation}\label{sec:difftpt}
% To solve this limitation, we develop a diverse data augmentation using diffusion models to generate data with richer visual appearance variation. 
% As shown in Fig.~\ref{fig:2} (a), we employ Stable Diffusion-V2 $\mathcal G(\mathbf{x}_\texttt{test})$, the text-to-image generative model that can produce new images conditioned on their natural language descriptions. For a given image $\mathbf{x}$, the latent representation $z_0$ is first obtained through a pre-trained encoder, along with CLIP text encoder for conditioning. A small amount of normally distributed Gaussian noise $\epsilon_t$ is then added to $z_0$ in finite steps $T$ until the representation converges to a normal distribution $z_T \sim N(0, I)$. Finally, the variation of the image is generated from the source dataset by denoising the normally distributed variable $z_T \sim N(0, I)$ with a limited steps $T$.
% By denoising the normally distributed variable $z_T \sim N(0, I)$ with a finite number of denoising steps $T$, the data distribution can be learned.

Diffusion-based data augmentation is presented to generate diverse and informative augmented images.
As shown in Fig.~\ref{fig:2} (a), 
from a given single test image $\mathbf{x}_\texttt{test}$, we first extract its latent features $z_0$ from the pre-trained CLIP encoder $f(\mathbf{x}_\texttt{test})$, and then use the stable diffusion as the decoder to generate augmented images. 
In specific, we employ Stable Diffusion-V2 as the generative model, which can generate new image $\mathcal G(g(\mathbf{t}), \mathbf{n})$ from the natural language descriptions $\mathbf{t}$.
Here, $\mathbf{n} \sim \mathcal{N}(0, I)$ denotes the sampled noise.
Because the labels are not available during test time tuning, we use the image encoder of CLIP model $f(\mathbf{x}_\texttt{test})$ as an alternative of $g(\mathbf{t})$. 
Thus, the augmented image can then be generated with
\vspace{-5pt}
\begin{equation}
\mathcal D_n(\mathbf{x}_\texttt{test}) = \mathcal G(f(\mathbf{x}_\texttt{test}), \mathbf{n}_n),
\vspace{-5pt}
\end{equation}
where $\mathcal D_n(\mathbf{x}_\texttt{test})$ denotes the $n$-th augmented image. 
Thanks to the ability of CLIP in aligning image and text, diffusion-based data augmentation is effective in generating diverse augmented images.
Please refer to the \emph{Suppl.} for visualizations of our diverse and informative augmentations.

%Therefore, after we obtain the latent features $z_0$, a small amount of normally distributed Gaussian noise $\epsilon_t$ is then added to $z_0$ in finite steps $T$ until the representation converges to a normal distribution $z_T \sim N(0, I)$. Finally, the variation of the image is generated from the source dataset by denoising the normally distributed variable $z_T \sim N(0, I)$ with a limited number of steps $T$. The diversely generated new data can be created by repeatedly querying the conditional generative model. Formally, such process can be expressed as
% In practice, we can sample variables from $z_T \sim N(0, I)$ to reconstruct the latent feature of a single test image $z_0$ and then decode by a pretrained decoder $\mathbf{x}_\texttt{test}=\mathcal{G}(z_0)$. 

Taking the augmented images $\{\mathcal D_n(\mathbf{x}_\texttt{test})\}$ into account, the test-time prompt tuning in Eq.~(\ref{tpt}) can be modified as,
\vspace{-8pt}
\begin{equation}
\tilde{\boldsymbol{p}}_{\boldsymbol{v}}=\frac{1}{\rho_H N} \sum_{n=1}^N \!\mathbbm{1}\!\left[\mathbf{H}\left( \boldsymbol{p}_i\right) \leq \tau\right] p_{\boldsymbol{n}}\left(y \!\mid \!\mathcal{D}_n\left(\mathbf{x}_{\texttt {test}}\right)\right). 
\label{difftpt}
\vspace{-3pt}
\end{equation}
Moreover, we incorporate augmented data by both the method in~\cite{shu2022test} and diffusion-based one to take advantage of their complementary merits. 
Then, we can still use Eq.~(\ref{tpt0}) to learn the adaptive prompt for the test sample $\mathbf{x}$.

%In this way, new images with richer visual appearance variation are created by denoising the latent variable $z_T$ conditioned on their representations. Formally, Eq.~(\ref{tpt}) can be rewritten as 
%\vspace{-5pt}
%\begin{equation}
%\tilde{\boldsymbol{p}}_{\boldsymbol{v}}=\frac{1}{\rho_H N} \sum_{i=1}^N \!\mathbbm{1}\!\left[\mathbf{H}\left(p_i\right) \leq \tau\right] p_{\boldsymbol{n}}\left(y \!\mid \!\mathcal{G}_i\left(\mathbf{x}_{\texttt {test}}\right)\right). 
%\label{difftpt}
%\vspace{-3pt}
%\end{equation}
%In particular, we incorporate augmented data by both conventional method and pre-trained stable diffusion to take advantage of the complementary strengths of the two data sources. The balance of the synthetic data and standard augment views is discussed in Sec.~\ref{sec:ab}. Accordingly, this approach provides an effective solution for the natural distributional variation problem of TPT in zero-shot generalization. 

\subsection{Filtration with Cosine Similarity}\label{sec:cosine}
\begin{figure}[t]
    % \vspace{-7pt}
	\begin{center}
		\includegraphics[width=\linewidth]{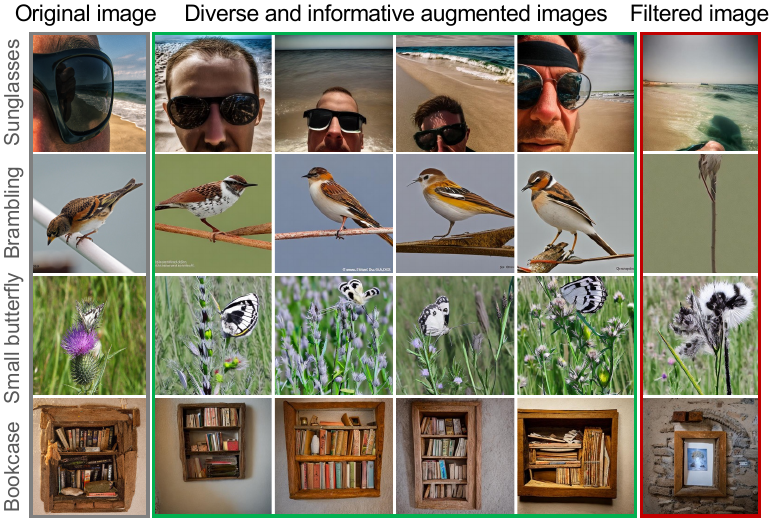}
	\end{center}
	\vspace{-15pt}
	\captionsetup{font=small}
    	\caption{\textbf{Visualization} of the diverse and informative diffusion-based augmented images and the filtered image by cosine similarity. }
	\vspace{-17pt}
	\label{fig:3.4}
\end{figure}

Albeit diffusion-based data augmentation is effective in generating diverse augmented images, some spurious augmentations may be introduced (see Fig.~\ref{fig:3.4}), resulting in low data fidelity and collapsing prompt tuning performance.  
%
%Though Eq.~(\ref{difftpt}) solves one aspect of the problem, we found there are still many unconfident yet correct predictions that are discarded (see Fig.~\ref{fig:1} (a)), resulting in low data fidelity. 
%
%Additionally, the pre-trained diffusion model enables more diverse and informative samples in the test set. 
%
Thus, it is necessary to balance the data diversity and prediction fidelity of the augmented images.

To tackle this issue, we introduce a cosine similarity based filteration approach.
In specific, we calculate the cosine similarity between the test sample $\mathbf{x}_{\texttt{test}}$ and each augmented image $\mathcal{D}_n(\mathbf{x}_{\texttt{test}})$. 
Then, we introduce a mask $\mathcal{M}$ to identify the augmented images with higher similarity than $\varepsilon$, \ie, $\mathcal{M}_n= \left(\cos\left(\mathcal{D}_n\left( \mathbf{x}_{\texttt {test}} \right), \mathbf{x}_\texttt{test}\right)\right)>\varepsilon$. 
We note that $\varepsilon$ is the threshold parameter resulting a $\rho_C$ percentage on the augmented images.
Taking both diffusion-based data augmentation and cosine similarity-based filteration into account, the test-time prompt tuning in Eq.~(\ref{tpt}) can be further modified as,
\vspace{-6pt}
\begin{equation}
\tilde{\boldsymbol{p}}_{\boldsymbol{v}}=\frac{1}{\rho_H \rho_C N} \sum_{i=1}^N \!\mathbbm{1}\!\left[\mathbf{H}\left(p_i\right) \leq \tau\right] \cdot \!\mathbbm{1}\!\left[\mathcal{M}_n\right] 
\boldsymbol{p}_{\boldsymbol{n}}\left(y \!\mid \!\mathcal{D}_n\left(\mathbf{x}_{\texttt {test}}\right)\right),
\vspace{-2pt}
\label{difftpt2}
\end{equation}
%where $\mathcal{M}_n= \left(\cos\left(\mathcal{D}_n\left( \mathbf{x}_{\texttt {test}} \right), \mathbf{x}_\texttt{test}\right)\right)>\tau^{\prime}$, $\tau^{\prime}$ is the threshold parameter resulting a $\rho_C$ percentage on the augmented images. 
Thus, we can obtain a large number of augmented samples with richer visual appearance variation while preserving key semantics to optimize the prompt during test-time.

\section{Experiments}
% \vspace{-2pt}
\subsection{Experimental Setup}
% \vspace{-4pt}
\noindent{\textbf{Implementation Details.}} 
Our method is implemented on one NVIDIA Tesla V$100$ GPU and $32$GB of memory. Following~\cite{shu2022test}, the initialized prompt is set to a hand-crafted default form, ``a photo of a'', and the corresponding $4$ tokens are optimized based on a single test image. We augment each test image to produce $63$ new images by Stable Diffusion-V2 in addition to the original one, and $64$ new images by different augment views~\cite{shu2022test}. The prompt is optimized with $4$ steps during test phase, where Adam is the optimizer. The hyper-parameter of initial learning rate, $\rho_{H}$, and $\rho_{C}$, are set to $0.005$, $0.3$, and $0.8$, respectively.

\noindent{\textbf{Datasets.}} 
We evaluate our method in two $\mathcal{S}$cenarios, \ie, $\mathcal{S}_1$: \texttt{Natural Distribution Shifts} and $\mathcal{S}_2$: \texttt{Cross-Datasets Generalization}. Following~\cite{shu2022test}, for $\mathcal{S}_1$, we use four datasets which are the out-of-distribution (OOD) data for \textbf{ImageNet}~\cite{deng2009imagenet}, \ie, \textbf{ImageNet-V2}~\cite{recht2019imagenet}, \textbf{ImageNet-A}~\cite{hendrycks2021natural}, \textbf{ImageNet-R}~\cite{hendrycks2021many}, and \textbf{ImageNet-Sketch}~\cite{wang2019learning} to investigate the robustness of our method to natural distribution shifts as these datasets differ in image style, data domains, \etc. For $\mathcal{S}_2$, $10$ datasets various in different species of plants or animals, scenes, textures, food, transportation, human actions, satellite images, and general objects are adopted in our experiments, \ie, \textbf{Flower102}~\cite{nilsback2008automated}, \textbf{OxfordPets}~\cite{parkhi2012cats}, \textbf{SUN397}~\cite{xiao2010sun}, \textbf{DTD}~\cite{cimpoi2014describing}, \textbf{Food101}~\cite{bossard2014food}, \textbf{StanfordCars}~\cite{krause20133d}, \textbf{Aircraft}~\cite{maji2013fine}, \textbf{UCF101}~\cite{soomro2012ucf101}, \textbf{EuroSAT}~\cite{helber2019eurosat}, and \textbf{Caltech101}~\cite{fei2004learning}. In particular, to investigate the effectiveness of our method with regard to cross-datasets generalization, ImageNet is used as a comprehensive source dataset, and the other $10$ datasets are used as target datasets for evaluation. In our experiments, we note that 1,000 test images are randomly selected from all the classes to evaluate all the methods.

% In particular, two different settings are adopted to investigate the effectiveness of our method with regard to cross-datasets generalization, \ie, 1) $\mathcal{S}_2{^{\prime}}$, where ImageNet is used as a comprehensive source dataset, and the other $10$ datasets are used as target datasets for evaluation, 2) $\mathcal{S}_2{^{\prime\prime}}$, a more challenging scenario where the source data for few-shot prompt tuning is consistent with the target dataset and without overlap of categories between the source and target pairs. We note that we randomly select $1000$ images among all classes of each dataset for testing. The testing results of the entire datasets are provided in Appendix.

% \vspace{-4pt}
\noindent{\textbf{Baselines.}}~To evaluate our proposed method, we adopt three groups of methods, \textbf{a)} TPT~\cite{shu2022test}, a state-of-the-art test-time prompt tuning method that is optimized upon multiple augmented views, \textbf{b)} the classical few-shot prompt tuning methods for CLIP, \ie, CoOp~\cite{zhou2022learning}, a few-shot prompt tuning baseline that tunes a fixed prompt on each downstream dataset, and CoCoOp~\cite{zhou2022conditional}, a improved few-shot prompt tuning baseline that generate input-conditional prompts by lightweight neural network, as well as \textbf{c)} two kinds of zero-shot CLIP, one with the ensemble of $80$ specially created prompts~\cite{radford2021learning}. and the other with the default prompt ``a picture of a.'' Following these works~\cite{zhou2022learning,zhou2022conditional,shu2022test}, all the baselines are trained on ImageNet with $16$-shot and $4$ learnable prompt tokens, and finally tested on OOD benchmarks.

\begin{table*}[t]
\renewcommand{\arraystretch}{1.3}
	\caption{\textbf{Top 1 accuracy} $\%$ of state-of-the-art baselines under $\mathcal{S}_1$, where \textbf{ImageNet-Sk.} indicates the ImageNet-Sketch dataset, \textbf{OOD Avg.} indicates the OOD average results. {\cellcolor{mygray}{$bs.$}} indicates the baseline of each group, \ie, CLIP-RN50 / CLIP-ViT-B-16, CoOp, and CoCoOp. The arrow ${\color{ForestGreen}\uparrow}$ and ${\color{red}\downarrow}$ indicate \textbf{improvements} and \textbf{decrements} compared with {\cellcolor{mygray}{$bs.$}}. Detailed analyses are provided in Sec.~\ref{sec:acc}.}
	\vspace{-4pt}
	\label{tab:1}
        
	\fontsize{8.5}{8.5}\selectfont
	\centering
	\begin{tabular}{l cc cc cc cc cc cc cc}
\toprule

\textbf{Method}
&\multicolumn{2}{c}{\textbf{~~ImageNet~~}}
&\multicolumn{2}{c}{\textbf{~~ImageNet-A~~}}
&\multicolumn{2}{c}{\textbf{~ImageNet-V2~}} 
&\multicolumn{2}{c}{\textbf{~~ImageNet-R~~}}
&\multicolumn{2}{c}{\textbf{~ImageNet-Sk.~}} 
&\multicolumn{2}{c}{\textbf{~~~~Average~~~~}} 
&\multicolumn{2}{c}{\textbf{~~OOD Avg.~~}}\\
\cmidrule(r){1-1}  \cmidrule{2-3} \cmidrule(lr){4-5} \cmidrule(lr){6-7} \cmidrule(lr){8-9}
\cmidrule(lr){10-11} \cmidrule(lr){12-13} \cmidrule(lr){14-15}

{{\cellcolor{mygray}CLIP-RN50}} 
&\multicolumn{2}{|c}{{\cellcolor{mygray}{$58.10$\stdvno{$bs.$}}}}
&\multicolumn{2}{c}{{\cellcolor{mygray}{$22.81$\stdvno{$bs.$}} }}
&\multicolumn{2}{c}{{\cellcolor{mygray}{$53.00$\stdvno{$bs.$}} }}
&\multicolumn{2}{c}{{\cellcolor{mygray}{$53.90$\stdvno{$bs.$}}}}
&\multicolumn{2}{c}{{\cellcolor{mygray}{$33.50$\stdvno{$bs.$}} }}
&\multicolumn{2}{|c}{{\cellcolor{mygray}{$42.26$\stdvno{$bs.$}} }}
&\multicolumn{2}{|c}{{\cellcolor{mygray}{$40.80$\stdvno{$bs.$}}}} \\

{\cellcolor{mypink2}Ensemble }
&\multicolumn{2}{|c}{{\cellcolor{mypink2}$59.90$\stdvu{$1.80$} }}
&\multicolumn{2}{c}{{\cellcolor{mypink2}$24.12$\stdvu{$1.31$} } }
&\multicolumn{2}{c}{{\cellcolor{mypink2}$53.50$\stdvu{$0.50$}} }
&\multicolumn{2}{c}{{\cellcolor{mypink2}$58.00$\stdvu{$4.10$}}}
&\multicolumn{2}{c}{{\cellcolor{mypink2}$35.20$\stdvu{$1.70$}}} 
&\multicolumn{2}{|c}{{\cellcolor{mypink2}$46.14$\stdvu{$3.88$}} }
&\multicolumn{2}{|c}{{\cellcolor{mypink2}$42.70$\stdvu{$1.90$}}} \\

{\cellcolor{mypink2}TPT }
&\multicolumn{2}{|c}{{\cellcolor{mypink2}$59.40$\stdvu{$1.30$} }}
&\multicolumn{2}{c}{{\cellcolor{mypink2}$27.34$\stdvu{$4.53$} } }
&\multicolumn{2}{c}{{\cellcolor{mypink2}$55.20$\stdvu{$2.20$}} }
&\multicolumn{2}{c}{{\cellcolor{mypink2}$56.80$\stdvu{$2.90$}}}
&\multicolumn{2}{c}{{\cellcolor{mypink2}$34.50$\stdvu{$1.00$}} }
&\multicolumn{2}{|c}{{\cellcolor{mypink2}$46.65$\stdvu{$4.39$}} }
&\multicolumn{2}{|c}{{\cellcolor{mypink2}$43.46$\stdvu{$2.66$}}} \\

{{\cellcolor{mypink2}\textbf{DiffTPT}}} 
&\multicolumn{2}{|c}{{\cellcolor{mypink2}$\textbf{60.80}$\stdvu{$\underline{{2.70}}$}}}
&\multicolumn{2}{c}{{\cellcolor{mypink2}$\textbf{31.06}$\stdvu{$\underline{{8.25}}$}}}
&\multicolumn{2}{c}{{\cellcolor{mypink2}$\textbf{55.80}$\stdvu{$\underline{{2.80}}$}}}
&\multicolumn{2}{c}{{\cellcolor{mypink2}$\textbf{58.80}$\stdvu{$\underline{{4.90}}$}}}
&\multicolumn{2}{c}{{\cellcolor{mypink2}$\textbf{37.10}$\stdvu{$\underline{{3.60}}$}}}
&\multicolumn{2}{|c}{{\cellcolor{mypink2}$\textbf{48.71}$\stdvu{$\underline{{6.45}}$}}} 
&\multicolumn{2}{|c}{{\cellcolor{mypink2}$\textbf{45.69}$\stdvu{$\underline{{4.89}}$}}} \\

{\cellcolor{mypink1}CoOp }
&\multicolumn{2}{|c}{{\cellcolor{mypink1}$63.30$\stdvno{$bs.$}}}
&\multicolumn{2}{c}{{\cellcolor{mypink1}$24.52$\stdvno{$bs.$}} }
&\multicolumn{2}{c}{{\cellcolor{mypink1}$57.90$\stdvno{$bs.$}} }
&\multicolumn{2}{c}{{\cellcolor{mypink1}$55.10$\stdvno{$bs.$}}}
&\multicolumn{2}{c}{{\cellcolor{mypink1}$34.80$\stdvno{$bs.$}} }
&\multicolumn{2}{|c}{{\cellcolor{mypink1}$47.12$\stdvno{$bs.$}} }
&\multicolumn{2}{|c}{{\cellcolor{mypink1}$43.08$\stdvno{$bs.$}}} \\

{\cellcolor{mypink1}TPT\&CoOp }
&\multicolumn{2}{|c}{{\cellcolor{mypink1}$63.70$\stdvu{$0.40$} }}
&\multicolumn{2}{c}{{\cellcolor{mypink1}$29.75$\stdvu{$5.23$} } }
&\multicolumn{2}{c}{{\cellcolor{mypink1}$60.90$\stdvu{$3.00$}} }
&\multicolumn{2}{c}{{\cellcolor{mypink1}$57.80$\stdvu{$2.70$}}}
&\multicolumn{2}{c}{{\cellcolor{mypink1}$36.50$\stdvu{$1.70$}} }
&\multicolumn{2}{|c}{{\cellcolor{mypink1}$49.73$\stdvu{$2.61$}} }
&\multicolumn{2}{|c}{{\cellcolor{mypink1}$46.24$\stdvu{$3.16$}}} \\

{{\cellcolor{mypink1}\textbf{DiffTPT\&CoOp}}} 
&\multicolumn{2}{|c}{{\cellcolor{mypink1}$\textbf{64.70}$\stdvu{$\underline{{1.40}}$}}}
&\multicolumn{2}{c}{{\cellcolor{mypink1}$\textbf{32.96}$\stdvu{$\underline{{8.44}}$}}}
&\multicolumn{2}{c}{{\cellcolor{mypink1}$\textbf{61.70}$\stdvu{$\underline{{3.80}}$}}}
&\multicolumn{2}{c}{{\cellcolor{mypink1}$\textbf{58.20}$\stdvu{$\underline{{3.10}}$}}}
&\multicolumn{2}{c}{{\cellcolor{mypink1}$\textbf{36.80}$\stdvu{$\underline{{2.00}}$}}}
&\multicolumn{2}{|c}{{\cellcolor{mypink1}$\textbf{50.87}$\stdvu{$\underline{{3.75}}$}}} 
&\multicolumn{2}{|c}{{\cellcolor{mypink1}$\textbf{47.42}$\stdvu{$\underline{{4.34}}$}}} \\

{\cellcolor{mypink}CoCoOp }
&\multicolumn{2}{|c}{{\cellcolor{mypink}$61.50$\stdvno{$bs.$} }}
&\multicolumn{2}{c}{{\cellcolor{mypink}$25.73$\stdvno{$bs.$} } }
&\multicolumn{2}{c}{{\cellcolor{mypink}$54.80$\stdvno{$bs.$}} }
&\multicolumn{2}{c}{{\cellcolor{mypink}$56.00$\stdvno{$bs.$}}}
&\multicolumn{2}{c}{{\cellcolor{mypink}$35.50$\stdvno{$bs.$}} }
&\multicolumn{2}{|c}{{\cellcolor{mypink}$46.71$\stdvno{$bs.$}} }
&\multicolumn{2}{|c}{{\cellcolor{mypink}$43.01$\stdvno{$bs.$}}} \\

{\cellcolor{mypink}TPT\&CoCoOp }
&\multicolumn{2}{|c}{{\cellcolor{mypink}$62.40$\stdvu{$0.90$} }}
&\multicolumn{2}{c}{{\cellcolor{mypink}$26.43$\stdvu{$0.70$} } }
&\multicolumn{2}{c}{{\cellcolor{mypink}$56.10$\stdvu{$1.30$}} }
&\multicolumn{2}{c}{{\cellcolor{mypink}$56.50$\stdvu{$0.50$}}}
&\multicolumn{2}{c}{{\cellcolor{mypink}$35.60$\stdvu{$0.10$}} }
&\multicolumn{2}{|c}{{\cellcolor{mypink}$47.41$\stdvu{$0.70$}} }
&\multicolumn{2}{|c}{{\cellcolor{mypink}$43.66$\stdvu{$0.65$}} }\\

{{\cellcolor{mypink}\textbf{DiffTPT\&CoCoOp}}} 
&\multicolumn{2}{|c}{{\cellcolor{mypink}$\textbf{63.50}$\stdvu{$\underline{{2.00}}$}}}
&\multicolumn{2}{c}{{\cellcolor{mypink}$\textbf{30.45}$\stdvu{$\underline{{0.72}}$}}}
&\multicolumn{2}{c}{{\cellcolor{mypink}$\textbf{57.70}$\stdvu{$\underline{{2.90}}$}}}
% &\multicolumn{2}{c}{{\cellcolor{mypink}$\textbf{59.20}$\stdvu{$\underline{{3.20}}$}}}
&\multicolumn{2}{c}{{\cellcolor{mypink}$\textbf{58.50}$\stdvu{$\underline{{2.50}}$}}}
&\multicolumn{2}{c}{{\cellcolor{mypink}$\textbf{37.90}$\stdvu{$\underline{{2.40}}$}}}
% &\multicolumn{2}{|c}{{\cellcolor{mypink}$\textbf{49.75}$\stdvu{$\underline{{3.04}}$}}} 
&\multicolumn{2}{|c}{{\cellcolor{mypink}$\textbf{49.61}$\stdvu{$\underline{{2.90}}$}}} 
% &\multicolumn{2}{|c}{{\cellcolor{mypink}$\textbf{46.31}$\stdvu{$\underline{{3.30}}$}}} \\\hline\hline
&\multicolumn{2}{|c}{{\cellcolor{mypink}$\textbf{46.14}$\stdvu{$\underline{{3.13}}$}}} \\\hline\hline

{{\cellcolor{mygray}{CLIP-ViT-B/16} }}
&\multicolumn{2}{|c}{{\cellcolor{mygray}{$67.30$\stdvno{$bs.$} }}}
&\multicolumn{2}{c}{{\cellcolor{mygray}{$47.14$\stdvno{$bs.$} }}} 
&\multicolumn{2}{c}{{\cellcolor{mygray}{$59.90$\stdvno{$bs.$}}}} 
&\multicolumn{2}{c}{{\cellcolor{mygray}{$71.20$\stdvno{$bs.$}}}}
&\multicolumn{2}{c}{{\cellcolor{mygray}{$43.00$\stdvno{$bs.$}} }}
&\multicolumn{2}{|c}{{\cellcolor{mygray}{$57.71$\stdvno{$bs.$}}}} 
&\multicolumn{2}{|c}{{\cellcolor{mygray}{$55.31$\stdvno{$bs.$}}}} \\

{\cellcolor{mypink2}Ensemble }
&\multicolumn{2}{|c}{{\cellcolor{mypink2}$68.50$\stdvu{$1.20$} }}
&\multicolumn{2}{c}{{\cellcolor{mypink2}$48.44$\stdvu{$1.30$} } }
&\multicolumn{2}{c}{{\cellcolor{mypink2}$62.70$\stdvu{$2.80$}} }
&\multicolumn{2}{c}{{\cellcolor{mypink2}$73.50$\stdvu{$2.30$}}}
&\multicolumn{2}{c}{{\cellcolor{mypink2}$45.50$\stdvu{$2.20$}} }
&\multicolumn{2}{|c}{{\cellcolor{mypink2}$59.73$\stdvu{$2.02$}} }
&\multicolumn{2}{|c}{{\cellcolor{mypink2}$57.53$\stdvu{$2.22$}} }\\

{\cellcolor{mypink2}TPT }
&\multicolumn{2}{|c}{{\cellcolor{mypink2}$69.70$\stdvu{$2.40$} }}
&\multicolumn{2}{c}{{\cellcolor{mypink2}$53.67$\stdvu{$6.53$} } }
&\multicolumn{2}{c}{{\cellcolor{mypink2}$64.30$\stdvu{$4.40$}} }
&\multicolumn{2}{c}{{\cellcolor{mypink2}$73.90$\stdvu{$2.70$}}}
&\multicolumn{2}{c}{{\cellcolor{mypink2}$46.40$\stdvu{$3.40$}} }
&\multicolumn{2}{|c}{{\cellcolor{mypink2}$61.59$\stdvu{$3.88$}} }
&\multicolumn{2}{|c}{{\cellcolor{mypink2}$59.57$\stdvu{$4.26$}}} \\

{{\cellcolor{mypink2}\textbf{DiffTPT}}} 
&\multicolumn{2}{|c}{{\cellcolor{mypink2}$\textbf{70.30}$\stdvu{$\underline{{3.00}}$}}}
&\multicolumn{2}{c}{{\cellcolor{mypink2}$\textbf{55.68}$\stdvu{$\underline{{8.54}}$}}}
&\multicolumn{2}{c}{{\cellcolor{mypink2}$\textbf{65.10}$\stdvu{$\underline{{5.20}}$}}}
&\multicolumn{2}{c}{{\cellcolor{mypink2}$\textbf{75.00}$\stdvu{$\underline{{3.80}}$}}}
&\multicolumn{2}{c}{{\cellcolor{mypink2}$\textbf{46.80}$\stdvu{$\underline{{3.80}}$}}}
&\multicolumn{2}{|c}{{\cellcolor{mypink2}$\textbf{62.28}$\stdvu{$\underline{{4.57}}$}}} 
&\multicolumn{2}{|c}{{\cellcolor{mypink2}$\textbf{60.52}$\stdvu{$\underline{{5.21}}$}}} \\

{\cellcolor{mypink1}CoOp }
&\multicolumn{2}{|c}{{\cellcolor{mypink1}$72.30$\stdvno{$bs.$} }}
&\multicolumn{2}{c}{{\cellcolor{mypink1}$49.25$\stdvno{$bs.$} } }
&\multicolumn{2}{c}{{\cellcolor{mypink1}$65.70$\stdvno{$bs.$}} }
&\multicolumn{2}{c}{{\cellcolor{mypink1}$71.50$\stdvno{$bs.$}}}
&\multicolumn{2}{c}{{\cellcolor{mypink1}$47.60$\stdvno{$bs.$}} }
&\multicolumn{2}{|c}{{\cellcolor{mypink1}$61.27$\stdvno{$bs.$}} }
&\multicolumn{2}{|c}{{\cellcolor{mypink1}$58.51$\stdvno{$bs.$}}} \\

{\cellcolor{mypink1}TPT\&CoOp }
&\multicolumn{2}{|c}{{\cellcolor{mypink1}$73.30$\stdvu{$1.00$} }}
&\multicolumn{2}{c}{{\cellcolor{mypink1}$56.88$\stdvu{$7.63$} } }
&\multicolumn{2}{c}{{\cellcolor{mypink1}$66.60$\stdvu{$0.90$}} }
&\multicolumn{2}{c}{{\cellcolor{mypink1}$73.80$\stdvu{$2.30$}}}
&\multicolumn{2}{c}{{\cellcolor{mypink1}$49.40$\stdvu{$1.80$}} }
&\multicolumn{2}{|c}{{\cellcolor{mypink1}$64.00$\stdvu{$2.73$}} }
&\multicolumn{2}{|c}{{\cellcolor{mypink1}$61.67$\stdvu{$3.16$}}} \\

{{\cellcolor{mypink1}\textbf{DiffTPT\&CoOp}}} 
&\multicolumn{2}{|c}{{\cellcolor{mypink1}$\textbf{75.00}$\stdvu{$\underline{{2.70}}$}}}
&\multicolumn{2}{c}{{\cellcolor{mypink1}$\textbf{58.09}$\stdvu{$\underline{{8.84}}$}}}
&\multicolumn{2}{c}{{\cellcolor{mypink1}$\textbf{66.80}$\stdvu{$\underline{{1.10}}$}}}
&\multicolumn{2}{c}{{\cellcolor{mypink1}$\textbf{73.90}$\stdvu{$\underline{{2.40}}$}}}
&\multicolumn{2}{c}{{\cellcolor{mypink1}$\textbf{49.50}$\stdvu{$\underline{{1.90}}$}}}
&\multicolumn{2}{|c}{{\cellcolor{mypink1}$\textbf{64.12}$\stdvu{$\underline{{2.85}}$}}} 
&\multicolumn{2}{|c}{{\cellcolor{mypink1}$\textbf{61.97}$\stdvu{$\underline{{3.46}}$}}} \\

{\cellcolor{mypink}CoCoOp }
&\multicolumn{2}{|c}{{\cellcolor{mypink}$71.40$\stdvno{$bs.$} }}
&\multicolumn{2}{c}{{\cellcolor{mypink}$50.05$\stdvno{$bs.$} } }
&\multicolumn{2}{c}{{\cellcolor{mypink}$63.80$\stdvno{$bs.$}} }
&\multicolumn{2}{c}{{\cellcolor{mypink}$73.10$\stdvno{$bs.$}}}
&\multicolumn{2}{c}{{\cellcolor{mypink}$46.70$\stdvno{$bs.$}} }
&\multicolumn{2}{|c}{{\cellcolor{mypink}$61.01$\stdvno{$bs.$}} }
&\multicolumn{2}{|c}{{\cellcolor{mypink}$58.41$\stdvno{$bs.$}}} \\

{\cellcolor{mypink}TPT\&CoCoOp }
&\multicolumn{2}{|c}{{\cellcolor{mypink}$67.30$\stdvd{$4.10$} }}
&\multicolumn{2}{c}{{\cellcolor{mypink}$50.25$\stdvu{$0.20$} } }
&\multicolumn{2}{c}{{\cellcolor{mypink}$62.30$\stdvd{$1.50$}} }
&\multicolumn{2}{c}{{\cellcolor{mypink}$73.90$\stdvu{$0.80$}}}
&\multicolumn{2}{c}{{\cellcolor{mypink}$47.10$\stdvu{$0.40$}} }
&\multicolumn{2}{|c}{{\cellcolor{mypink}$60.17$\stdvd{$0.84$}} }
&\multicolumn{2}{|c}{{\cellcolor{mypink}$58.39$\stdvd{$0.02$}} }\\

{{\cellcolor{mypink}\textbf{DiffTPT\&CoCoOp}}} 
&\multicolumn{2}{|c}{{\cellcolor{mypink}$\textbf{69.30}$\stdvd{$\underline{{2.10}}$}}}
&\multicolumn{2}{c}{{\cellcolor{mypink}$\textbf{52.56}$\stdvu{$\underline{{2.51}}$}}}
&\multicolumn{2}{c}{{\cellcolor{mypink}$\textbf{63.20}$\stdvd{$\underline{{0.60}}$}}}
&\multicolumn{2}{c}{{\cellcolor{mypink}$\textbf{75.30}$\stdvu{$\underline{{2.20}}$}}}
&\multicolumn{2}{c}{{\cellcolor{mypink}$\textbf{47.50}$\stdvu{$\underline{{0.80}}$}}}
&\multicolumn{2}{|c}{{\cellcolor{mypink}$\textbf{61.57}$\stdvu{$\underline{{0.56}}$}}} 
&\multicolumn{2}{|c}{{\cellcolor{mypink}$\textbf{59.64}$\stdvu{$\underline{{1.23}}$}}} \\
\bottomrule
\end{tabular}
\label{tab:1}
\vspace{-10pt}
\end{table*}

\begin{figure*}[!t]
\renewcommand{\arraystretch}{1.3}
	\makeatletter\def\@captype{table}\makeatother\caption{\textbf{Top 1 accuracy} $\%$ of state-of-the-art baselines under $\mathcal{S}_2$, where \textbf{Avg}. indicates average accuracies of the \texttt{Cross-Datasets Generalization}. The arrow ${\color{ForestGreen}\uparrow}$ and ${\color{red}\downarrow}$ indicate \textbf{improvements} and \textbf{decrements} of our method against the CLIP method, \ie, CLIP-RN50 and CLIP-ViT-B/16. Detailed analyses are provided in Sec.~\ref{sec:acc}.}
	% \vspace{-7pt}
	\label{tab:2}\centering
 \begin{minipage}[c]{\textwidth}
        
	\fontsize{8.6}{8.6}\selectfont
	\centering
	\begin{tabular}{l cc cc cc cc cc  }
\toprule
\centering
\textbf{Method}

&\multicolumn{2}{c}{\textbf{~~~~~~~Flower~\cite{nilsback2008automated}~~~~~~}}
&\multicolumn{2}{c}{\textbf{~~~~~~DTD~\cite{cimpoi2014describing}~~~~~~}}
&\multicolumn{2}{c}{\textbf{~~~~~~Pets~\cite{parkhi2012cats}~~~~~~}} 
&\multicolumn{2}{c}{\textbf{~~~~~~Cars~\cite{krause20133d}~~~~~~}}
&\multicolumn{2}{c}{\textbf{~~~~~~UCF101~\cite{soomro2012ucf101}~~~~~~}} \\
% &\multicolumn{2}{c}{\textbf{Caltech11}} 
% &\multicolumn{2}{c}{\textbf{Food101}}
% &\multicolumn{2}{c}{\textbf{SUN397}}
% &\multicolumn{2}{c}{\textbf{Aircraft}}
% &\multicolumn{2}{c}{\textbf{EuroSAT}}
% &\multicolumn{2}{c}{\textbf{Avg.}} \\

\cmidrule(r){1-1}  \cmidrule(lr){2-3} \cmidrule(lr){4-5} \cmidrule(lr){6-7} \cmidrule(lr){8-9} \cmidrule(lr){10-11} 
% \cmidrule(lr){12-13} \cmidrule(lr){14-15} \cmidrule(lr){16-17}\cmidrule(lr){18-19}\cmidrule(lr){20-21} \cmidrule(lr){22-23} 

{{\cellcolor{mygray}CLIP-RN50}} 
&\multicolumn{2}{|c}{{\cellcolor{mygray}{$62.45$\stdvno{$bs.$}}}}
&\multicolumn{2}{c}{{\cellcolor{mygray}{$39.65$\stdvno{$bs.$}} }}
&\multicolumn{2}{c}{{\cellcolor{mygray}{$80.50$\stdvno{$bs.$}} }}
&\multicolumn{2}{c}{{\cellcolor{mygray}{$57.48$\stdvno{$bs.$}}}}
&\multicolumn{2}{c}{{\cellcolor{mygray}{$56.73$\stdvno{$bs.$}} }}\\

Ensemble 
&\multicolumn{2}{|c}{$63.14$}
&\multicolumn{2}{c}{$41.68$} 
&\multicolumn{2}{c}{$80.79$} 
&\multicolumn{2}{c}{$58.33$}
&\multicolumn{2}{c}{$55.74$} \\

CoOp${\color{magenta}_{2022}}$~\cite{zhou2022learning}
&\multicolumn{2}{|c}{$62.25$}
&\multicolumn{2}{c}{$37.33$} 
&\multicolumn{2}{c}{$86.00$} 
&\multicolumn{2}{c}{$56.29$}
&\multicolumn{2}{c}{$59.01$}\\

CoCoOp${\color{magenta}_{2022}}$~\cite{zhou2022conditional}
&\multicolumn{2}{|c}{$63.53$}
&\multicolumn{2}{c}{$38.49$} 
&\multicolumn{2}{c}{$86.29$} 
&\multicolumn{2}{c}{$55.70$}
&\multicolumn{2}{c}{$60.40$} \\

TPT${\color{magenta}_{2022}}$~\cite{shu2022test} 
&\multicolumn{2}{|c}{$62.25$\stdvd{${{0.20}}$}}
&\multicolumn{2}{c}{$40.04$\stdvu{${{0.39}}$}} 
&\multicolumn{2}{c}{$82.82$\stdvu{${{2.32}}$}} 
&\multicolumn{2}{c}{$60.54$\stdvu{${{3.06}}$}}
&\multicolumn{2}{c}{$60.79$\stdvu{${{4.06}}$}} \\

{{\cellcolor{mypink}\textbf{DiffTPT}}} 
&\multicolumn{2}{|c}{{\cellcolor{mypink}$\textbf{63.53}$\stdvu{$\underline{{1.08}}$}}}
&\multicolumn{2}{c}{{\cellcolor{mypink}$\textbf{40.72}$\stdvu{$\underline{{1.07}}$}}}
&\multicolumn{2}{c}{{\cellcolor{mypink}$\textbf{83.40}$\stdvu{$\underline{{3.35}}$}}}
&\multicolumn{2}{c}{{\cellcolor{mypink}$\textbf{60.71}$\stdvu{$\underline{{3.23}}$}}}
&\multicolumn{2}{c}{{\cellcolor{mypink}$\textbf{62.67}$\stdvu{$\underline{{5.94}}$}}}\\

% \stdvu{$\underline{\textbf{0}}$}

\bottomrule
\end{tabular}
% \vspace{-15pt}
\end{minipage}

%==================================================================2

\centering
 \begin{minipage}[c]{\textwidth}\centering
        \centering
	\fontsize{8.6}{8.6}\selectfont
	\centering
	\begin{tabular}{l cc cc cc cc cc cc  }
 % \centering
\toprule
\centering
\textbf{Method}

&\multicolumn{2}{c}{\textbf{~Caltech11~\cite{fei2004learning}~}} 
&\multicolumn{2}{c}{\textbf{~Food101~\cite{bossard2014food}~}}
&\multicolumn{2}{c}{\textbf{SUN397~\cite{xiao2010sun}~}}
&\multicolumn{2}{c}{\textbf{~Aircraft~\cite{maji2013fine}~}}
&\multicolumn{2}{c}{\textbf{~EuroSAT~\cite{helber2019eurosat}~}}
&\multicolumn{2}{c}{\textbf{Avg.}} \\

\cmidrule(r){1-1} 
\cmidrule(lr){2-3} \cmidrule(lr){4-5} \cmidrule(lr){6-7} \cmidrule(lr){8-9} \cmidrule(lr){10-11} \cmidrule(lr){12-13} 

{{\cellcolor{mygray}CLIP-RN50}} 
&\multicolumn{2}{|c}{{\cellcolor{mygray}{$81.58$\stdvno{$bs.$}} }}
&\multicolumn{2}{c}{{\cellcolor{mygray}{$74.85$\stdvno{$bs.$}}}} 
&\multicolumn{2}{c}{{\cellcolor{mygray}{$57.43$\stdvno{$bs.$}}}}
&\multicolumn{2}{c}{{\cellcolor{mygray}{$16.20$\stdvno{$bs.$}} }}
&\multicolumn{2}{c}{{\cellcolor{mygray}{$24.30$\stdvno{$bs.$}} }}
&\multicolumn{2}{|c}{{\cellcolor{mygray}{$55.12$\stdvno{$bs.$}}}} 

\\

Ensemble
&\multicolumn{2}{|c}{$83.68$} 
&\multicolumn{2}{c}{$74.95$}
&\multicolumn{2}{c}{$59.53$}
&\multicolumn{2}{c}{$17.40$} 
&\multicolumn{2}{c}{$27.69$} 
&\multicolumn{2}{|c}{$56.29$} \\

CoOp${\color{magenta}_{2022}}$~\cite{zhou2022learning}
&\multicolumn{2}{|c}{$82.38$} 
&\multicolumn{2}{c}{$78.81$}
&\multicolumn{2}{c}{$57.18$}
&\multicolumn{2}{c}{$15.40$} 
&\multicolumn{2}{c}{$26.99$} 
&\multicolumn{2}{|c}{$56.16$} \\

CoCoOp${\color{magenta}_{2022}}$~\cite{zhou2022conditional}
&\multicolumn{2}{|c}{$83.38$} 
&\multicolumn{2}{c}{$77.43$}
&\multicolumn{2}{c}{$59.28$}
&\multicolumn{2}{c}{$15.70$} 
&\multicolumn{2}{c}{$27.39$} 
&\multicolumn{2}{|c}{$56.76$} \\

TPT${\color{magenta}_{2022}}$~\cite{shu2022test}
&\multicolumn{2}{|c}{$84.58$\stdvu{${{3.00}}$}} 
&\multicolumn{2}{c}{$77.23$\stdvu{${{2.38}}$}}
&\multicolumn{2}{c}{$61.80$\stdvu{${{4.37}}$}}
&\multicolumn{2}{c}{$17.50$\stdvu{${{1.30}}$}} 
&\multicolumn{2}{c}{$22.21$\stdvd{${{2.09}}$}} 
&\multicolumn{2}{|c}{$56.98$\stdvu{${{1.86}}$}} \\

{{\cellcolor{mypink}\textbf{DiffTPT}}} 
&\multicolumn{2}{|c}{{\cellcolor{mypink}$\textbf{86.89}$\stdvu{$\underline{{5.31}}$}}} 
&\multicolumn{2}{c}{{\cellcolor{mypink}$\textbf{79.21}$\stdvu{$\underline{{4.36}}$}}}
&\multicolumn{2}{c}{{\cellcolor{mypink}$\textbf{62.72}$\stdvu{$\underline{{5.29}}$}}}
&\multicolumn{2}{c}{{\cellcolor{mypink}$\textbf{17.60}$\stdvu{$\underline{{1.40}}$}}}
&\multicolumn{2}{c}{{\cellcolor{mypink}$\textbf{41.04}$\stdvu{$\underline{{16.74}}$}}} 
&\multicolumn{2}{|c}{{\cellcolor{mypink}$\textbf{59.85}$\stdvu{$\underline{{4.68}}$}}} \\

\bottomrule
\end{tabular}
% \vspace{-15pt}
\end{minipage} 

%========================================================================3
 \begin{minipage}[c]{\textwidth}
        
	\fontsize{8.6}{8.6}\selectfont
	\centering
	\begin{tabular}{l cc cc cc cc cc  }
\toprule
\centering
\textbf{Method}

&\multicolumn{2}{c}{\textbf{~~~~~~~Flower~\cite{nilsback2008automated}~~~~~~}}
&\multicolumn{2}{c}{\textbf{~~~~~~DTD~\cite{cimpoi2014describing}~~~~~~}}
&\multicolumn{2}{c}{\textbf{~~~~~~Pets~\cite{parkhi2012cats}~~~~~~}} 
&\multicolumn{2}{c}{\textbf{~~~~~~Cars~\cite{krause20133d}~~~~~~}}
&\multicolumn{2}{c}{\textbf{~~~~~~UCF101~\cite{soomro2012ucf101}~~~~~~}} \\

\cmidrule(r){1-1}  \cmidrule(lr){2-3} \cmidrule(lr){4-5} \cmidrule(lr){6-7} \cmidrule(lr){8-9} \cmidrule(lr){10-11}

{{\cellcolor{mygray}CLIP-ViT-B/16}} 
&\multicolumn{2}{|c}{{\cellcolor{mygray}{$67.94$\stdvno{$bs.$} }}}
&\multicolumn{2}{c}{{\cellcolor{mygray}{$44.10$\stdvno{$bs.$} }}} 
&\multicolumn{2}{c}{{\cellcolor{mygray}{$85.71$\stdvno{$bs.$}}}} 
&\multicolumn{2}{c}{{\cellcolor{mygray}{$66.58$\stdvno{$bs.$}}}}
&\multicolumn{2}{c}{{\cellcolor{mygray}{$63.37$\stdvno{$bs.$}} }}\\

Ensemble 
&\multicolumn{2}{|c}{$67.65$}
&\multicolumn{2}{c}{$44.87$} 
&\multicolumn{2}{c}{$86.20$} 
&\multicolumn{2}{c}{$67.60$}
&\multicolumn{2}{c}{$64.36$}  \\

CoOp${\color{magenta}_{2022}}$~\cite{zhou2022learning} 
&\multicolumn{2}{|c}{$66.08$}
&\multicolumn{2}{c}{$42.17$} 
&\multicolumn{2}{c}{$89.00$} 
&\multicolumn{2}{c}{$63.44$}
&\multicolumn{2}{c}{$66.04$} \\

CoCoOp${\color{magenta}_{2022}}$~\cite{zhou2022conditional} 
&\multicolumn{2}{|c}{$70.88$}
&\multicolumn{2}{c}{$44.78$} 
&\multicolumn{2}{c}{$88.71$} 
&\multicolumn{2}{c}{$65.22$}
&\multicolumn{2}{c}{$68.42$}  \\

TPT${\color{magenta}_{2022}}$~\cite{shu2022test} 
&\multicolumn{2}{|c}{$69.31$\stdvu{${{1.37}}$}}
&\multicolumn{2}{c}{$46.23$\stdvu{${{2.13}}$}} 
&\multicolumn{2}{c}{$86.49$\stdvu{${{0.78}}$}} 
&\multicolumn{2}{c}{$66.50$\stdvd{${{0.08}}$}}
&\multicolumn{2}{c}{$66.44$\stdvu{${{3.07}}$}} \\

{{\cellcolor{mypink}\textbf{DiffTPT}}} 
&\multicolumn{2}{|c}{{\cellcolor{mypink}$\textbf{70.10}$\stdvu{$\underline{{2.16}}$}}}
&\multicolumn{2}{c}{{\cellcolor{mypink}$\textbf{47.00}$\stdvu{$\underline{{2.90}}$}}}
&\multicolumn{2}{c}{{\cellcolor{mypink}$\textbf{88.22}$\stdvu{$\underline{{2.51}}$}}}
&\multicolumn{2}{c}{{\cellcolor{mypink}$\textbf{67.01}$\stdvu{$\underline{{0.43}}$}}}
&\multicolumn{2}{c}{{\cellcolor{mypink}$\textbf{68.22}$\stdvu{$\underline{{4.85}}$}}}\\

\bottomrule
\end{tabular}
% \vspace{-15pt}
\end{minipage}

%==================================================================4

\centering
 \begin{minipage}[c]{\textwidth}\centering
        \centering
	\fontsize{8.6}{8.6}\selectfont
	\centering
	\begin{tabular}{l cc cc cc cc cc cc  }
 % \centering
\toprule
\centering
\textbf{Method}

&\multicolumn{2}{c}{\textbf{~Caltech11~\cite{fei2004learning}~}} 
&\multicolumn{2}{c}{\textbf{~Food101~\cite{bossard2014food}~}}
&\multicolumn{2}{c}{\textbf{SUN397~\cite{xiao2010sun}~}}
&\multicolumn{2}{c}{\textbf{~Aircraft~\cite{maji2013fine}~}}
&\multicolumn{2}{c}{\textbf{~EuroSAT~\cite{helber2019eurosat}~}}
&\multicolumn{2}{c}{\textbf{Avg.}} \\

\cmidrule(r){1-1} 
\cmidrule(lr){2-3} \cmidrule(lr){4-5} \cmidrule(lr){6-7} \cmidrule(lr){8-9} \cmidrule(lr){10-11} \cmidrule(lr){12-13} 

{{\cellcolor{mygray}CLIP-ViT-B/16}} 
&\multicolumn{2}{|c}{{\cellcolor{mygray}{$90.29$\stdvno{$bs.$}}}} 
&\multicolumn{2}{c}{{\cellcolor{mygray}{$85.05$\stdvno{$bs.$}}}}
&\multicolumn{2}{c}{{\cellcolor{mygray}{$61.88$\stdvno{$bs.$}}}}
&\multicolumn{2}{c}{{\cellcolor{mygray}{$24.70$\stdvno{$bs.$}} }}
&\multicolumn{2}{c}{{\cellcolor{mygray}{$40.64$\stdvno{$bs.$}}}} 
&\multicolumn{2}{|c}{{\cellcolor{mygray}{$63.03$\stdvno{$bs.$}}}}\\

Ensemble
&\multicolumn{2}{|c}{$90.89$} 
&\multicolumn{2}{c}{$85.35$}
&\multicolumn{2}{c}{$64.65$}
&\multicolumn{2}{c}{$24.40$} 
&\multicolumn{2}{c}{$47.01$} 
&\multicolumn{2}{|c}{$64.30$} \\

CoOp${\color{magenta}_{2022}}$~\cite{zhou2022learning}
&\multicolumn{2}{|c}{$91.69$} 
&\multicolumn{2}{c}{$85.15$}
&\multicolumn{2}{c}{$61.54$}
&\multicolumn{2}{c}{$18.00$} 
&\multicolumn{2}{c}{$35.36$} 
&\multicolumn{2}{|c}{$61.85$} \\

CoCoOp${\color{magenta}_{2022}}$~\cite{zhou2022conditional}
&\multicolumn{2}{|c}{$92.49$} 
&\multicolumn{2}{c}{$86.53$}
&\multicolumn{2}{c}{$64.65$}
&\multicolumn{2}{c}{$24.20$} 
&\multicolumn{2}{c}{$46.22$} 
&\multicolumn{2}{|c}{$65.21$} \\

TPT${\color{magenta}_{2022}}$~\cite{shu2022test}
&\multicolumn{2}{|c}{$92.49$\stdvu{${{2.20}}$}} 
&\multicolumn{2}{c}{$86.93$\stdvu{${{1.88}}$}}
&\multicolumn{2}{c}{$63.48$\stdvu{${{1.60}}$}}
&\multicolumn{2}{c}{$24.90$\stdvu{${{0.20}}$}} 
&\multicolumn{2}{c}{$37.15$\stdvd{${{3.49}}$}} 
&\multicolumn{2}{|c}{$63.99$\stdvu{${{0.96}}$}} \\

{{\cellcolor{mypink}\textbf{DiffTPT}}} 
&\multicolumn{2}{|c}{{\cellcolor{mypink}$\textbf{92.49}$\stdvu{$\underline{{2.20}}$}}} 
&\multicolumn{2}{c}{{\cellcolor{mypink}$\textbf{87.23}$\stdvu{$\underline{{2.18}}$}}}
&\multicolumn{2}{c}{{\cellcolor{mypink}$\textbf{65.74}$\stdvu{$\underline{{3.86}}$}}}
&\multicolumn{2}{c}{{\cellcolor{mypink}$\textbf{25.60}$\stdvu{$\underline{{0.90}}$}}}
&\multicolumn{2}{c}{{\cellcolor{mypink}$\textbf{43.13}$\stdvu{$\underline{{3.49}}$}}} 
&\multicolumn{2}{|c}{{\cellcolor{mypink}$\textbf{65.47}$\stdvu{$\underline{{2.44}}$}}} \\

\bottomrule
\end{tabular}
% \vspace{-13pt}
\end{minipage} 

\end{figure*}

\vspace{-2pt}
\subsection{Comparison with State-of-the-arts}\label{sec:acc}
\vspace{-2pt}
% In this section, we provide a quantitative comparison both on $\mathcal{S}_1$ and $\mathcal{S}_2$. %of our proposed method with various baselines.

\noindent{\textbf{Natural Distribution Shifts.}}~Table~\ref{tab:1} summarizes the evaluation of competing methods under $\mathcal{S}$cenario 1 and different backbones, \ie, ResNet-50 and ViT-B/16, where ensemble refers to zero-shot CLIP performance under an ensemble of $80$ hand-crafted prompts and CLIP refers to the zero-shot CLIP performance by default prompt ``a photo of a''. ``---\&CoOp'' and ``---\&CoCoOp'' indicate applying the test-time prompt tuning method to the CoOp~\cite{zhou2022learning} or CoCoOp~\cite{zhou2022conditional}, which are fine-tuned with $16$ shot training data per category on ImageNet. As can be seen from this table, DiffTPT outperforms all the other methods on the five datasets. 
Applying DiffTPT to prompts learned by CoOp~\cite{zhou2022learning} or CoCoOp~\cite{zhou2022conditional} can both improve the accuracy of their in-domain \textbf{ImageNet} data as well as their generalization ability to OOD data, \ie, based on a ResNet-50, the average of in-domain \textbf{ImageNet} data is improved from $47.12$ to $\textbf{50.87}$ and $46.71$ to $\textbf{49.61}$, respectively, and the average for OOD generation is improved from $43.08$ to $\textbf{47.42}$ and $43.01$ to $\textbf{46.14}$, respectively. Similar improvements of our method are also obtained on ViT-B/16, \ie, $61.27$ $\rightarrow$ $\textbf{64.12}$, and $61.01$ $\rightarrow$ $\textbf{61.57}$ on ResNet-50, $58.51$ $\rightarrow$ $\textbf{61.97}$, and $58.41$ $\rightarrow$ $\textbf{59.64}$ on ViT-B/16.
Compared with TPT, TPT\&CoOp, and TPT\&CoCoOp, our proposed methods, DiffTPT, DiffTPT\&CoOp, and DiffTPT\&CoCoOp, increase classification accuracy in-domain \textbf{ImageNet} data from $46.65$, $49.73$, $47.41$ to $\textbf{48.71}$, $\textbf{50.87}$, $\textbf{49.61}$, respectively.
Since the TPT-based method uses random resized crops to augment test image, making it limited in generalization ability. 
In particular, we discover that DiffTPT significantly improves the generalization test of OOD data. This supports our conclusion that DiffTPT increases the \textit{\textbf{data diversity}} of the test samples while maintaining semantic consistency (\ie, \textit{\textbf{prediction fidelity}}), resulting in better robustness. 
On the contrary, the performance of the few-shot prompt tuning method differs significantly across the five datasets; \eg, the accuracy gains are prominent on the \textbf{ImageNet} validation set and \textbf{ImageNet-V2}, while on the datasets with more significant distribution shift, even prompt tuning methods become brittle compared to hand-crafted prompts.

\begin{figure*}[t]
    % \vspace{-7pt}
	\begin{center}
		\includegraphics[width=0.98\linewidth]{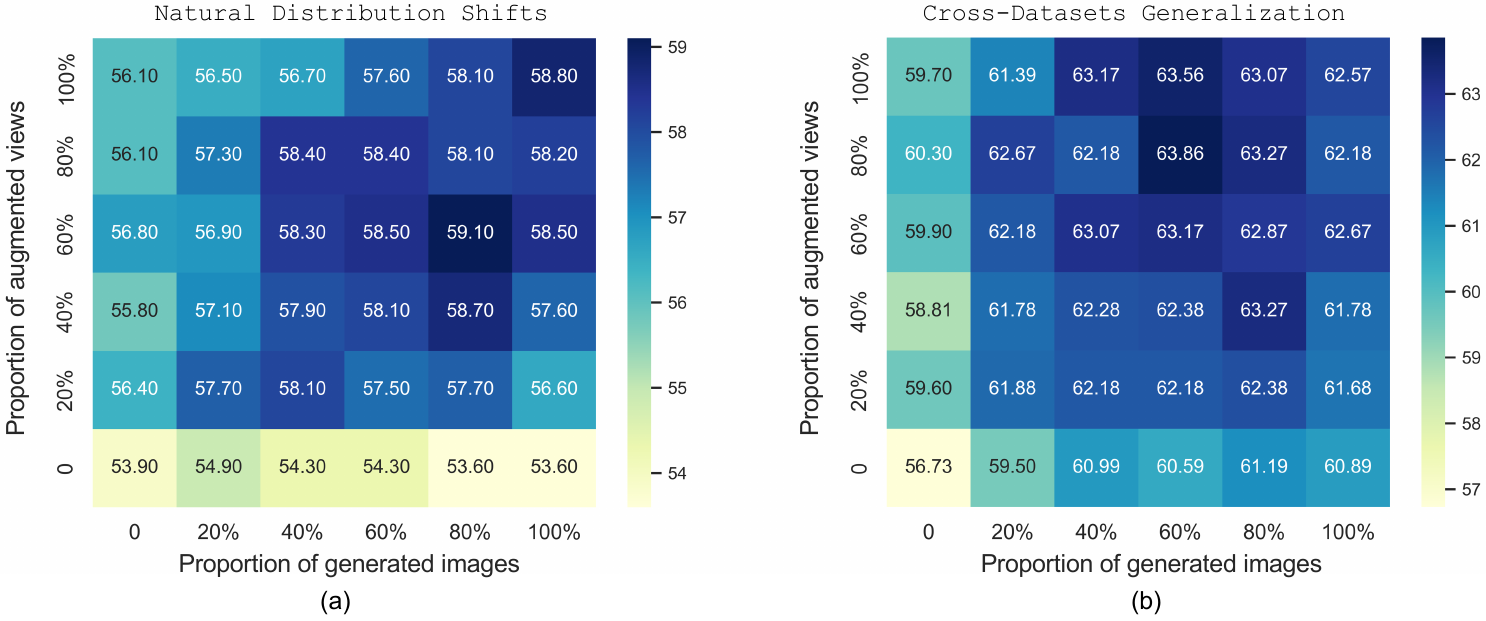}
            % \put(-408,2.7){ \small$M$}
            % \put(-238,2.2){ \small$c$}
            % \put(-73,3.4){ \small$\alpha$}
	\end{center}
	\vspace{-18pt}
	\captionsetup{font=small}
    	\caption{\textbf{Variation of the top 1 accuracy} versus the varied proportion of the \textit{standard} augmented views and the \textit{diffusion-based} augmented images  under (a) $\mathcal{S}_1$ and (b) $\mathcal{S}_2$.}
	\vspace{-6pt}
	\label{fig:3}
\end{figure*}

Naturally, CLIP yields the lowest results, and testing directly on the new dataset will be significantly affected by domain shift. Despite the fact that they benefit from the learnable prompts and the complement of TPT, CoOp~\cite{zhou2022learning} and CoCoOp~\cite{zhou2022conditional} perform better against CLIP; however, these methods require the training set and are not test-time prompt tuning methods. That is, they did not consider the zero-shot generalization in real world scenarios, resulting in less effective performance. 
The results support our primal hypothesis that augmenting test data with diverse synthetic data can improve zero-shot generalization performance. 

% against the top-performing method

\noindent{\textbf{Cross-Datasets Generalization.}}~To investigate how our proposed method and baselines generalize from ImageNet to the 10 fine-grained datasets, we record the quantitative performance of various baselines of $\mathcal{S}_2$ in Table~\ref{tab:2}. TPT~\cite{shu2022test} works in a zero-shot manner, while CoOp~\cite{zhou2022learning} and CoCoOp~\cite{zhou2022conditional} are tuned on \textbf{ImageNet} using $16$-shot training data per category. From this table, since the distribution of these fine-grained datasets varies widely, these methods perform differently on each dataset. 
However, our method still achieves the best performance, \ie, increasing the \textbf{Avg.} accuracy from $55.12$ to $\textbf{59.85}$, and from $63.03$ to $\textbf{65.47}$.
It is worth noting that our method still achieves $5.1$\% and $2.3$\% performance gain against TPT~\cite{shu2022test} on the backbone of both ResNet$50$ and Vit-B/$16$. 
This indicates that among all competing methods, even without training data, our method is robust to natural distributional variations and significantly outperforms those few-shot prompt-tuning methods, \ie, CoOp~\cite{zhou2022learning} and CoCoOp~\cite{zhou2022conditional}.

\subsection{Ablation Studies}\label{sec:ab}
\vspace{-2pt}
\noindent{\textbf{Balancing Synthetic Data \textit{vs.}~Standard Augmentation.}}~
Since our method absorbs the complementary merits of both standard augmentation~\cite{shu2022test} and diffusion-based one, it is necessary to investigate how these two methods help train classifiers. To this end, we evaluate the average performance of ResNet$50$ in the two scenarios, \ie, $\mathcal{S}_1$ and $\mathcal{S}_2$. For better visualization, we plot the mixed combinations of various ratios in Fig.~\ref{fig:3}, where the abscissa and ordinate represent the proportions of synthetic data obtained by diffusion-based augmentation and data obtained by standard augmentation, respectively. In the matrix of this figure, each element $\mathcal{M}_{ij}$ represents the classification performance of DiffTPT on $i$\% of synthetic data and $j$\% of standard augmented data. As shown in Fig.~\ref{fig:3} (a), one can observe an improvement in accuracy of the \texttt{Natural Distribution Shifts} as the size of the standard augmented data grows, while maintaining a constant amount of synthetic data. However, similar results are more obvious when the proportion of synthetic data is increased while keeping the proportion of standard augmented data fixed. Overall, increasing the amount of synthetic data leads to an improved performance in $\mathcal{S}_1$. In Fig.~\ref{fig:3} (b), we show the performance of the classifier for $\mathcal{S}_2$, \ie, \texttt{Cross-Datasets Generalization}. We observe that, while keeping the amount of synthetic data fixed, the effectiveness of the classifier increases significantly as the proportion of standard augmented data increases. These conclusions are supported by the results of most datasets, see the \textbf{\texttt{Suppl.}} for more details.

\begin{figure}[t]
    \vspace{-5pt}
	\begin{center}
		\includegraphics[width=\linewidth]{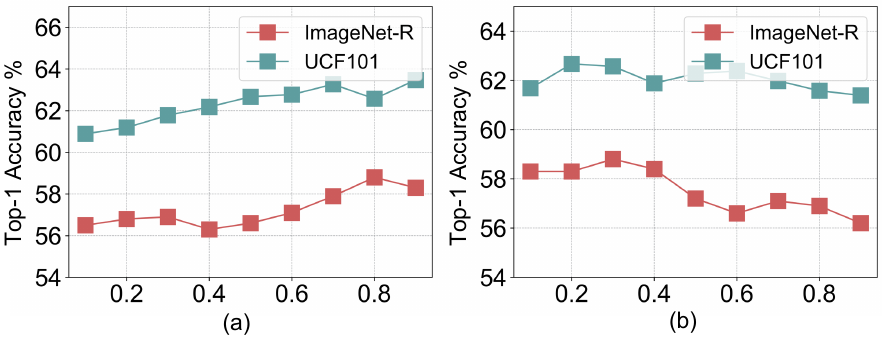}
            \put(-169,2.5){ \small$\rho_H$}
            \put(-48,2.8){ \small$\rho_C$}
	\end{center}
	\vspace{-17pt}
	\captionsetup{font=small}
	\caption{\small{Top-1 accuracy analysis of the \textbf{ratios $\rho_H$} and $\rho_C$ with regard to $\mathcal{S}_1$ and $\mathcal{S}_2$.}}
	\vspace{-14pt}
	\label{fig:4}
\end{figure}

\noindent{\textbf{Analysis of Ratio $\rho_H$ and $\rho_C$.}}~As mentioned in Sec.~\ref{sec:cosine}, $\rho_H$ and $\rho_C$ filter out the less informative ``noisy'' augmented view in standard augmentation by entropy and spurious augmentations in diffusion-based augmentation by cosine similarity. We examine the classification accuracy for various values of $\rho_H$ and $\rho_C$ for the two scenarios in Fig.~\ref{fig:4} to evaluate the diverse information that a good test augmentation should retain. It can be seen from Fig.~\ref{fig:4} (a) that as the value of $\rho_C$ increases, the classification accuracy gradually improves. 
This is because a high threshold will result in too much effective information being filtered out and reduce the effect of test set augmentation. However, from Fig.~\ref{fig:4} (b), we find that when $\rho_H > 0.5$, it will lead to a decrease in classification accuracy. In summary, when the value of $\rho_H$ is too large and $\rho_C$ is too small, the model goes to an extreme of overemphasizing the augmented data, giving rise to unsatisfactory performance. To sum up, our method achieves the highest results on $\rho_H = 0.3$ and $\rho_C = 0.8$. This means that only a small amount of data is filtered out, indicating the high fidelity of the synthetic data by diffusion-based augmentation.

\begin{figure}[t]
    \vspace{-10pt}
	\begin{center}
		\includegraphics[width=0.98\linewidth]{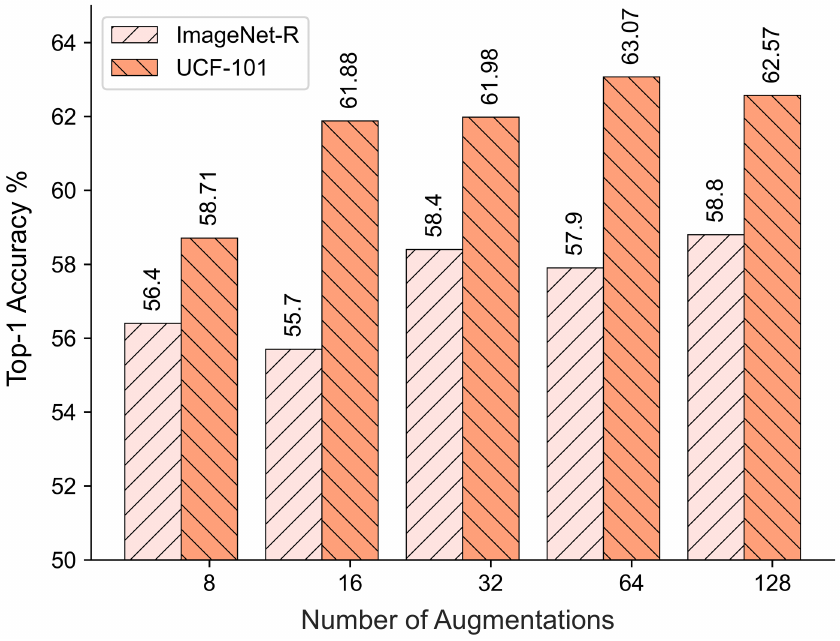}
	\end{center}
	\vspace{-18pt}
	\captionsetup{font=small}
    	\caption{\textbf{Ablation studies} on the size of augmented images with regard to $\mathcal{S}_1$ and $\mathcal{S}_2$.}
	\vspace{-2pt}
	\label{fig:5}
\end{figure}
\noindent{\textbf{Effect of the Generated Dataset Size.}}~The results on a variety of datasets in both two scenarios demonstrate that our approach is effective for all images regardless of their category and content (see Table~\ref{tab:1} and~\ref{tab:2}). Considering the effect of augmented data on model performance, we recorded the classification accuracy of different numbers of augmented samples in two scenarios in Fig.~\ref{fig:5}. It can be seen that as the number of augmented views increases, the accuracy gradually increases until reaching a plateau around $N = 64$. Additionally, even under the more complex scenario, \ie, $\mathcal{S}_2$, our method can still preserve similar results. 
Notably, even when $N = 8$, DiffTPT still brings more than $3.5$\% and $4.6$\% accuracy gain for zero-shot CLIP, respectively.
When $N = 128$, the performance of DiffTPT did not appear to be degraded in accuracy due to overemphasizing synthetic data. These positive experimental results support that our cosine similarity filteration is effective in preserving prediction fidelity between synthetic and original data.

\begin{figure}[t]
    \vspace{-4pt}
	\begin{center}
		\includegraphics[width=\linewidth]{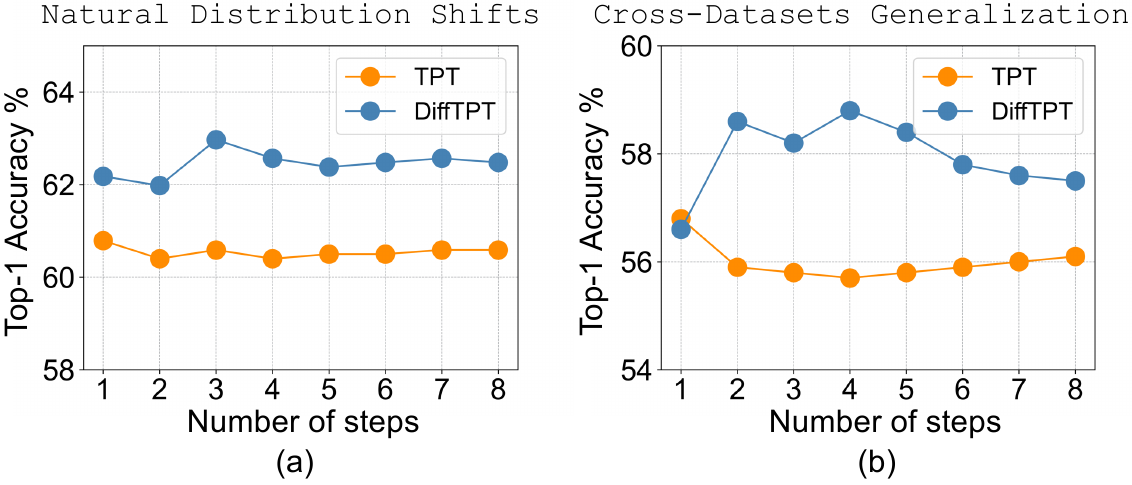}
	\end{center}
	\vspace{-19pt}
	\captionsetup{font=small}
    	\caption{\textbf{Ablation studies} on the \textit{steps of prompt updating} under $\mathcal{S}_1$ and $\mathcal{S}_2$.}
	\vspace{-20pt}
	\label{fig:6}
\end{figure}

\noindent{\textbf{Steps of Prompt Updating.}}~To assess the effect of prompt updating, we record the accuracy of different optimization steps under the two scenarios in Fig.~\ref{fig:6}. 
As can be seen from the figure, the accuracy can be improved from $56.6$ to $\textbf{58.8}$ by increasing the number of optimization steps from $1$ to $4$ on $\mathcal{S}_2$, while the performance will decrease slightly when the optimization is continued (\ie, optimization steps greater than $5$).
This shows that more optimization steps cannot bring gains to the classifier; on the contrary, a few updates can make the prompt learn more information about the test samples. For TPT~\cite{shu2022test}, more update steps did not make the classifier optimize better, while step $= 1$ achieved the best performance. Nevertheless, our method always outperforms TPT~\cite{shu2022test} in terms of the classification accuracies on the two scenarios. Additionally, since more update steps will increase the inference time, we set optimization step to $4$ as the default setting in our experiments.

\noindent{\textbf{Inference cost.}}~
% Yes, test-time prompt tuning generally is computationally expensive, \eg, it takes $6$s to inference $10$ test images for TPT and with standard SD it is $36$min but we also tried recent consistency model\footnote{Consistency Models, arXiv 2023.\label{consm}} that gives similar performance gain but with only \textbf{6}min. Additionally, our experiment with Memory Efficient Attention\footnote{https://www.photoroom.com/tech/stable-diffusion-100-percent-faster-with-memory-efficient-attention\label{mea}} leads to a further \textbf{100}\% gain in inference speed.
Albeit the original SD is time-consuming, \eg, it takes $6$s to inference $10$ test images for TPT and with standard SD it is $36$min, with the advancements in this field, faster SD models have emerged, \eg, ToMe~\cite{bolya2023token}, two-stage distillation~\cite{meng2023distillation}, and Consistency Model~\cite{song2023consistency}.
% \footnotemark[\value{Consistency Models, arXiv 2023.}]. 
The later only need $0.5$s for generating $10$ images, while the original SD needs $70$s. Additionally, there are a few strategies to boost the SD efficiency, \eg, TensorRT and Memory Efficient Attention\footnote{https://www.photoroom.com/tech/stable-diffusion-100-percent-faster-with-memory-efficient-attention}.
% \footnotemark[\value{https://www.photoroom.com/tech/stable-diffusion-100-percent-faster-with-memory-efficient-attention}]. 
These approaches lead to a further $25$\% and $100$\% gains in inference speed.

\vspace{-2pt}
\section{Conclusion}
\vspace{-2pt}
This work proposed a simple but effective test-time prompt tuning method,  based on a pre-trained diffusion model, DiffTPT, by incorporating diffusion-based augmentation with cosine similarity-based filteration. In specific, with the pre-trained diffusion model, diffusion-based augmentation can generate diverse but semantically consistent augmented images.
And cosine similarity-based filteration can enforce the prediction fidelity of the generated samples. Extensive experiments demonstrated the effectiveness of DiffTPT in various zero-shot generalization tasks, \eg, DiffTPT improves zero-shot accuracy by an average of $5.13$\% in comparison to the state-of-the-art test-time prompt-tuning method, TPT.

\noindent\textbf{Acknowledgements:}
This work was supported by the National Research Foundation, Singapore under its AI Singapore Programme (AISG Award No: AISG2-TC-2021-003), Agency for Science, Technology and Research (A*STAR) through its AME Programmatic Funding Scheme Under Project A20H4b0141, A*STAR Central Research Fund "A Secure and Privacy Preserving AI Platform for Digital Health”, and Agency for Science, Technology and Research (A*STAR) through its RIE2020 Health and Biomedical Sciences (HBMS) Industry Alignment Fund Pre-Positioning (IAF-PP) (grant no. H20C6a0032).

%%%%%%%%% REFERENCES
{\small
\bibliographystyle{ieee_fullname}
\bibliography{egbib}
}
\clearpage
\appendix

\setcounter{table}{0}
\renewcommand{\thetable}{A\arabic{table}}
\setcounter{figure}{0}
\renewcommand{\thefigure}{A\arabic{figure}}
\section*{Contents}
The following items are included in our supplementary material:
\begin{itemize}
  \item Accuracy \% of competing methods on the full dataset of $\mathcal{S}_1$ and $\mathcal{S}_2$ in Section~\ref{sec:A}.
  \item Proportion analysis of the two kinds of augmented images in Section~\ref{sec:C}.
  \item Visualization of the diverse and informative diffusion-based augmented images and the filtered image in Section~\ref{sec:B}.
  % \item Additional qualitative results for \texttt{In-Federation} and \texttt{Out-of-Federation} scenarios in Section~\ref{sec:A4}.
  
\end{itemize}

\section{Comparisons of the Full Dataset on $\mathcal{S}_1$ and $\mathcal{S}_2$}\label{sec:A}

We summarize the results of all competing methods with regard to $\mathcal{S}_1$ and $\mathcal{S}_2$ on the full dataset in Table~\ref{tab:A1}, also including different backbones \ie, ResNet-50 and ViT-B/16. The results in this table are completely consistent with those in Tables {\color{red}1} and {\color{red}2}, and our method DiffTPT outperforms the other methods on almost all the datasets. For example, on the \textbf{ImageNet-V2} dataset of $\mathcal{S}_1$, DiffTPT improves the performance of CLIP-RN50 from $51.41$ $\%$ to $\textbf{55.87}$ $\%$. However, the performance of TPT~\cite{shu2022test} is lower than that of CoCoOp, \ie, CoCoOp: $55.72$ $\%$, \textit{vs}. TPT: $55.07$ $\%$. Especially on the EuroSAT dataset, compared to the baselines CLIP-RN50 and CLIP-ViT-B/16, TPT~\cite{shu2022test} reduces the accuracy by $0.87$ $\%$ and $3.75$ $\%$, respectively. On the contrary, our method gets significant improvement, \ie, on the backbone of ResNet-50, TPT~\cite{shu2022test}: $22.56$ $\%$ \textit{vs}. DiffTPT: $\textbf{41.70}$ $\%$, and on the backbone of ViT-B/16, TPT~\cite{shu2022test}: $38.26$ $\%$ \textit{vs}. DiffTPT: $\textbf{45.20}$ $\%$. In addition, although the average results of TPT~\cite{shu2022test} on all datasets are improved in comparison to the baseline, \ie, ResNet-50: $52.98$ $\%$ $\rightarrow$ 
$\textbf{54.42}$ $\%$, ViT-B/16: $61.16$ $\%$ $\rightarrow$ 
$\textbf{62.21}$ $\%$, our proposed method DiffTPT achieved higher improvements, \ie, ResNet-50: $54.42$ $\%$ $\rightarrow$ $\textbf{57.52}$ $\%$, ViT-B/16: $62.21$ $\%$ $\rightarrow$ 
$\textbf{63.27}$ $\%$. These results on the full datasets show that among all competing methods, even without training data, our method is robust to adapt to distribution variations and significantly outperforms those few-shot prompt tuning methods, \eg, CoOp~\cite {zhou2022learning} and CoCoOp~\cite{zhou2022conditional}, and the state-of-the-art test-time prompt-tuning method, TPT~\cite{shu2022test}.

\begin{figure*}[!t]
\renewcommand{\arraystretch}{1.3}
 \makeatletter\def\@captype{table}\makeatother\caption{\textbf{Top 1 accuracy} $\%$ of the competing methods on the full dataset of $\mathcal{S}_1$ and $\mathcal{S}_2$. The arrow ${\color{ForestGreen}\uparrow}$ and ${\color{red}\downarrow}$ indicate \textbf{improvements} and \textbf{decrements} of our method against the CLIP method, \ie, CLIP-RN50 and CLIP-ViT-B/16. Detailed analyses are provided in Sec.~\ref{sec:A}.}
	% \vspace{-7pt}
	\label{tab:A1}\centering
 \begin{minipage}[c]{\textwidth}
        
	\fontsize{8.6}{8.6}\selectfont
	\centering
	\begin{tabular}{l cc cc cc cc cc}
\toprule
\centering
\textbf{Method}

&\multicolumn{2}{c}{\textbf{~~~~ImageNet-V2~\cite{recht2019imagenet}~~~}}
&\multicolumn{2}{c}{\textbf{~~~~~~~Flower~\cite{nilsback2008automated}~~~~~~}}
&\multicolumn{2}{c}{\textbf{~~~~~~DTD~\cite{cimpoi2014describing}~~~~~~}}
&\multicolumn{2}{c}{\textbf{~~~~~~Pets~\cite{parkhi2012cats}~~~~~~}} 
&\multicolumn{2}{c}{\textbf{~~~~~~Cars~\cite{krause20133d}~~~~~~}}\\

% &\multicolumn{2}{c}{\textbf{Caltech11}} 
% &\multicolumn{2}{c}{\textbf{Food101}}
% &\multicolumn{2}{c}{\textbf{SUN397}}
% &\multicolumn{2}{c}{\textbf{Aircraft}}
% &\multicolumn{2}{c}{\textbf{EuroSAT}}
% &\multicolumn{2}{c}{\textbf{Avg.}} \\

\cmidrule(r){1-1}  \cmidrule(lr){2-3} \cmidrule(lr){4-5} \cmidrule(lr){6-7} \cmidrule(lr){8-9} \cmidrule(lr){10-11} 
% \cmidrule(lr){12-13} \cmidrule(lr){14-15} \cmidrule(lr){16-17}\cmidrule(lr){18-19}\cmidrule(lr){20-21} \cmidrule(lr){22-23} 

{{\cellcolor{mygray}CLIP-RN50}} 
&\multicolumn{2}{|c|}{{\cellcolor{mygray}{$51.41$\stdvno{$bs.$}}}}
&\multicolumn{2}{c}{{\cellcolor{mygray}{$61.75$\stdvno{$bs.$}} }}
&\multicolumn{2}{c}{{\cellcolor{mygray}{$40.37$\stdvno{$bs.$}} }}
&\multicolumn{2}{c}{{\cellcolor{mygray}{$83.57$\stdvno{$bs.$}}}}
&\multicolumn{2}{c}{{\cellcolor{mygray}{$55.70$\stdvno{$bs.$}} }}\\

Ensemble 
&\multicolumn{2}{|c|}{$52.91$}
&\multicolumn{2}{c}{$62.77$} 
&\multicolumn{2}{c}{$40.37$} 
&\multicolumn{2}{c}{$82.97$}
&\multicolumn{2}{c}{$55.89$} \\

CoOp${\color{magenta}_{2022}}$~\cite{zhou2022learning}
&\multicolumn{2}{|c|}{$55.40$}
&\multicolumn{2}{c}{$61.55$} 
&\multicolumn{2}{c}{$37.29$} 
&\multicolumn{2}{c}{$87.00$}
&\multicolumn{2}{c}{$55.32$}\\

CoCoOp${\color{magenta}_{2022}}$~\cite{zhou2022conditional}
&\multicolumn{2}{|c|}{$55.72$}
&\multicolumn{2}{c}{$65.57$} 
&\multicolumn{2}{c}{$38.53$} 
&\multicolumn{2}{c}{$88.39$}
&\multicolumn{2}{c}{$56.22$} \\

TPT${\color{magenta}_{2022}}$~\cite{shu2022test} 
&\multicolumn{2}{|c|}{$55.07$\stdvu{${{3.66}}$}}
&\multicolumn{2}{c}{$62.57$\stdvu{${{0.82}}$}} 
&\multicolumn{2}{c}{$40.96$\stdvu{${{0.59}}$}} 
&\multicolumn{2}{c}{$84.44$\stdvu{${{0.87}}$}}
&\multicolumn{2}{c}{$58.56$\stdvu{${{2.86}}$}} \\

{{\cellcolor{mypink}\textbf{DiffTPT}}} 
&\multicolumn{2}{|c|}{{\cellcolor{mypink}$\textbf{55.87}$\stdvu{$\underline{{4.46}}$}}}
&\multicolumn{2}{c}{{\cellcolor{mypink}$\textbf{63.22}$\stdvu{$\underline{{1.47}}$}}}
&\multicolumn{2}{c}{{\cellcolor{mypink}$\textbf{41.31}$\stdvu{$\underline{{0.94}}$}}}
&\multicolumn{2}{c}{{\cellcolor{mypink}$\textbf{85.12}$\stdvu{$\underline{{1.55}}$}}}
&\multicolumn{2}{c}{{\cellcolor{mypink}$\textbf{59.33}$\stdvu{$\underline{{3.63}}$}}}\\

% \stdvu{$\underline{\textbf{0}}$}

\bottomrule
\end{tabular}
% \vspace{-15pt}
\end{minipage}

%==================================================================2

\centering
 \begin{minipage}[c]{\textwidth}\centering
        \centering
	\fontsize{8.6}{8.6}\selectfont
	\centering
	\begin{tabular}{l cc cc cc cc cc cc  }
 % \centering
\toprule
\centering
\textbf{Method}
&\multicolumn{2}{c}{\textbf{~~~~~~UCF101~\cite{soomro2012ucf101}~~~~~~}} 
&\multicolumn{2}{c}{\textbf{~~~~Caltech11~~~~\cite{fei2004learning}~}} 
% &\multicolumn{2}{c}{\textbf{~Food101~\cite{bossard2014food}~}}
% &\multicolumn{2}{c}{\textbf{SUN397~\cite{xiao2010sun}~}}
&\multicolumn{2}{c}{\textbf{~~~~Aircraft~~~~\cite{maji2013fine}~}}
&\multicolumn{2}{c}{\textbf{~~~EuroSAT~~~\cite{helber2019eurosat}~}}
&\multicolumn{2}{c}{\textbf{~~~Average~~~}} \\

\cmidrule(r){1-1} 
\cmidrule(lr){2-3} \cmidrule(lr){4-5} \cmidrule(lr){6-7} \cmidrule(lr){8-9} \cmidrule(lr){10-11} \cmidrule(lr){12-13} 

{{\cellcolor{mygray}CLIP-RN50}} 
&\multicolumn{2}{|c}{{\cellcolor{mygray}{$58.84$\stdvno{$bs.$}} }}
&\multicolumn{2}{c}{{\cellcolor{mygray}{$85.88$\stdvno{$bs.$}}}} 
&\multicolumn{2}{c}{{\cellcolor{mygray}{$15.66$\stdvno{$bs.$}}}}
&\multicolumn{2}{c}{{\cellcolor{mygray}{$23.69$\stdvno{$bs.$}} }}
&\multicolumn{2}{|c}{{\cellcolor{mygray}{$52.98$\stdvno{$bs.$}} }}

\\

Ensemble
&\multicolumn{2}{|c}{$59.48$} 
&\multicolumn{2}{c}{$87.26$}
&\multicolumn{2}{c}{$16.11$}
&\multicolumn{2}{c}{$25.79$} 
&\multicolumn{2}{|c}{$53.73$} \\

CoOp${\color{magenta}_{2022}}$~\cite{zhou2022learning}
&\multicolumn{2}{|c}{$59.05$} 
&\multicolumn{2}{c}{$86.53$}
&\multicolumn{2}{c}{$15.12$}
&\multicolumn{2}{c}{$26.20$} 
&\multicolumn{2}{|c}{$53.72$} \\

CoCoOp${\color{magenta}_{2022}}$~\cite{zhou2022conditional}
&\multicolumn{2}{|c}{$57.10$} 
&\multicolumn{2}{c}{$87.38$}
&\multicolumn{2}{c}{$14.61$}
&\multicolumn{2}{c}{$28.73$} 
&\multicolumn{2}{|c}{$54.69$} \\

TPT${\color{magenta}_{2022}}$~\cite{shu2022test}
&\multicolumn{2}{|c}{$60.77$\stdvu{${{1.93}}$}} 
&\multicolumn{2}{c}{$87.22$\stdvu{${{1.34}}$}}
&\multicolumn{2}{c}{$17.60$\stdvu{${{1.94}}$}}
&\multicolumn{2}{c}{$22.56$\stdvd{${{1.13}}$}} 
&\multicolumn{2}{|c}{$54.42$\stdvu{${{1.44}}$}} \\

{{\cellcolor{mypink}\textbf{DiffTPT}}} 
&\multicolumn{2}{|c}{{\cellcolor{mypink}$\textbf{63.20}$\stdvu{$\underline{{4.36}}$}}} 
&\multicolumn{2}{c}{{\cellcolor{mypink}$\textbf{89.70}$\stdvu{$\underline{{3.82}}$}}}
&\multicolumn{2}{c}{{\cellcolor{mypink}$\textbf{18.25}$\stdvu{$\underline{{2.59}}$}}}
&\multicolumn{2}{c}{{\cellcolor{mypink}$\textbf{41.70}$\stdvu{$\underline{{18.01}}$}}} 
&\multicolumn{2}{|c}{{\cellcolor{mypink}$\textbf{57.52}$\stdvu{$\underline{{4.54}}$}}} \\

\bottomrule
\end{tabular}
% \vspace{-15pt}
\end{minipage} 

%========================================================================3
 \begin{minipage}[c]{\textwidth}
        
	\fontsize{8.6}{8.6}\selectfont
	\centering
	\begin{tabular}{l cc cc cc cc cc  }
\toprule
\centering
\textbf{Method}
&\multicolumn{2}{c}{\textbf{~~~~ImageNet-V2~\cite{recht2019imagenet}~~~}}
&\multicolumn{2}{c}{\textbf{~~~~~~~Flower~\cite{nilsback2008automated}~~~~~~}}
&\multicolumn{2}{c}{\textbf{~~~~~~DTD~\cite{cimpoi2014describing}~~~~~~}}
&\multicolumn{2}{c}{\textbf{~~~~~~Pets~\cite{parkhi2012cats}~~~~~~}} 
&\multicolumn{2}{c}{\textbf{~~~~~~Cars~\cite{krause20133d}~~~~~~}}\\

\cmidrule(r){1-1}  \cmidrule(lr){2-3} \cmidrule(lr){4-5} \cmidrule(lr){6-7} \cmidrule(lr){8-9} \cmidrule(lr){10-11}

{{\cellcolor{mygray}CLIP-ViT-B/16}} 
&\multicolumn{2}{|c|}{{\cellcolor{mygray}{$60.86$\stdvno{$bs.$} }}}
&\multicolumn{2}{c}{{\cellcolor{mygray}{$67.44$\stdvno{$bs.$} }}} 
&\multicolumn{2}{c}{{\cellcolor{mygray}{$44.27$\stdvno{$bs.$}}}} 
&\multicolumn{2}{c}{{\cellcolor{mygray}{$88.25$\stdvno{$bs.$}}}}
&\multicolumn{2}{c}{{\cellcolor{mygray}{$65.48$\stdvno{$bs.$}} }}\\

Ensemble 
&\multicolumn{2}{|c|}{$61.88$}
&\multicolumn{2}{c}{$66.99$} 
&\multicolumn{2}{c}{$45.04$} 
&\multicolumn{2}{c}{$86.92$}
&\multicolumn{2}{c}{$66.11$}  \\

CoOp${\color{magenta}_{2022}}$~\cite{zhou2022learning} 
&\multicolumn{2}{|c|}{$64.20$}
&\multicolumn{2}{c}{$68.71$} 
&\multicolumn{2}{c}{$41.92$} 
&\multicolumn{2}{c}{$89.14$}
&\multicolumn{2}{c}{$64.51$} \\

CoCoOp${\color{magenta}_{2022}}$~\cite{zhou2022conditional} 
&\multicolumn{2}{|c|}{$64.07$}
&\multicolumn{2}{c}{$70.85$} 
&\multicolumn{2}{c}{$45.45$} 
&\multicolumn{2}{c}{$90.46$}
&\multicolumn{2}{c}{$64.90$}  \\

TPT${\color{magenta}_{2022}}$~\cite{shu2022test} 
&\multicolumn{2}{|c|}{$64.30$\stdvu{${{3.44}}$}}
&\multicolumn{2}{c}{$68.98$\stdvu{${{1.54}}$}} 
&\multicolumn{2}{c}{$47.70$\stdvu{${{3.43}}$}} 
&\multicolumn{2}{c}{$87.08$\stdvd{${{1.17}}$}}
&\multicolumn{2}{c}{$66.92$\stdvu{${{1.44}}$}} \\

{{\cellcolor{mypink}\textbf{DiffTPT}}} 
&\multicolumn{2}{|c|}{{\cellcolor{mypink}$\textbf{63.72}$\stdvu{$\underline{{2.86}}$}}}
&\multicolumn{2}{c}{{\cellcolor{mypink}$\textbf{69.47}$\stdvu{$\underline{{2.03}}$}}}
&\multicolumn{2}{c}{{\cellcolor{mypink}$\textbf{47.34}$\stdvu{$\underline{{3.07}}$}}}
&\multicolumn{2}{c}{{\cellcolor{mypink}$\textbf{87.95}$\stdvd{$\underline{{0.30}}$}}}
&\multicolumn{2}{c}{{\cellcolor{mypink}$\textbf{67.45}$\stdvu{$\underline{{1.97}}$}}}\\

\bottomrule
\end{tabular}
% \vspace{-15pt}
\end{minipage}

%==================================================================4

\centering
 \begin{minipage}[c]{\textwidth}\centering
        \centering
	\fontsize{8.6}{8.6}\selectfont
	\centering
	\begin{tabular}{l cc cc cc cc cc cc  }
 % \centering
\toprule
\centering
\textbf{Method}
&\multicolumn{2}{c}{\textbf{~~~~~~UCF101~\cite{soomro2012ucf101}~~~~~~}} 
&\multicolumn{2}{c}{\textbf{~~~~Caltech11~~~~\cite{fei2004learning}~}} 
% &\multicolumn{2}{c}{\textbf{~Food101~\cite{bossard2014food}~}}
% &\multicolumn{2}{c}{\textbf{SUN397~\cite{xiao2010sun}~}}
&\multicolumn{2}{c}{\textbf{~~~~Aircraft~~~~\cite{maji2013fine}~}}
&\multicolumn{2}{c}{\textbf{~~~EuroSAT~~~\cite{helber2019eurosat}~}}
&\multicolumn{2}{c}{\textbf{~~~Average~~~}} \\

\cmidrule(r){1-1} 
\cmidrule(lr){2-3} \cmidrule(lr){4-5} \cmidrule(lr){6-7} \cmidrule(lr){8-9} \cmidrule(lr){10-11} \cmidrule(lr){12-13} 

{{\cellcolor{mygray}CLIP-ViT-B/16}} 
&\multicolumn{2}{|c}{{\cellcolor{mygray}{$65.13$\stdvno{$bs.$}}}} 
&\multicolumn{2}{c}{{\cellcolor{mygray}{$93.35$\stdvno{$bs.$}}}}
&\multicolumn{2}{c}{{\cellcolor{mygray}{$23.67$\stdvno{$bs.$}}}}
&\multicolumn{2}{c}{{\cellcolor{mygray}{$42.01$\stdvno{$bs.$}} }}
&\multicolumn{2}{|c}{{\cellcolor{mygray}{$61.16$\stdvno{$bs.$}}}}\\

Ensemble
&\multicolumn{2}{|c}{$65.16$} 
&\multicolumn{2}{c}{$93.55$}
&\multicolumn{2}{c}{$23.22$}
&\multicolumn{2}{c}{$50.42$} 
&\multicolumn{2}{|c}{$62.14$} \\

CoOp${\color{magenta}_{2022}}$~\cite{zhou2022learning}
&\multicolumn{2}{|c}{$66.55$} 
&\multicolumn{2}{c}{$93.70$}
&\multicolumn{2}{c}{$18.47$}
&\multicolumn{2}{c}{$46.39$} 
&\multicolumn{2}{|c}{$61.51$} \\

CoCoOp${\color{magenta}_{2022}}$~\cite{zhou2022conditional}
&\multicolumn{2}{|c}{$68.44$} 
&\multicolumn{2}{c}{$93.79$}
&\multicolumn{2}{c}{$22.29$}
&\multicolumn{2}{c}{$39.23$} 
&\multicolumn{2}{|c}{$62.16$} \\

TPT${\color{magenta}_{2022}}$~\cite{shu2022test}
&\multicolumn{2}{|c}{$67.83$\stdvu{${{2.70}}$}} 
&\multicolumn{2}{c}{$94.08$\stdvu{${{0.73}}$}}
&\multicolumn{2}{c}{$24.73$\stdvu{${{1.06}}$}}
&\multicolumn{2}{c}{$38.26$\stdvd{${{3.75}}$}} 
&\multicolumn{2}{|c}{$62.21$\stdvu{${{1.05}}$}} \\

{{\cellcolor{mypink}\textbf{DiffTPT}}} 
&\multicolumn{2}{|c}{{\cellcolor{mypink}$\textbf{68.68}$\stdvu{$\underline{{3.55}}$}}} 
&\multicolumn{2}{c}{{\cellcolor{mypink}$\textbf{94.69}$\stdvu{$\underline{{1.34}}$}}}
&\multicolumn{2}{c}{{\cellcolor{mypink}$\textbf{24.96}$\stdvu{$\underline{{1.29}}$}}}
&\multicolumn{2}{c}{{\cellcolor{mypink}$\textbf{45.20}$\stdvu{$\underline{{3.19}}$}}}
&\multicolumn{2}{|c}{{\cellcolor{mypink}$\textbf{63.27}$\stdvu{$\underline{{2.11}}$}}} \\

\bottomrule
\end{tabular}
% \vspace{-13pt}
\end{minipage} 

\end{figure*}

\begin{figure*}[!t]
    % \vspace{-15pt}
	\begin{center}
		\includegraphics[width=0.4\linewidth]{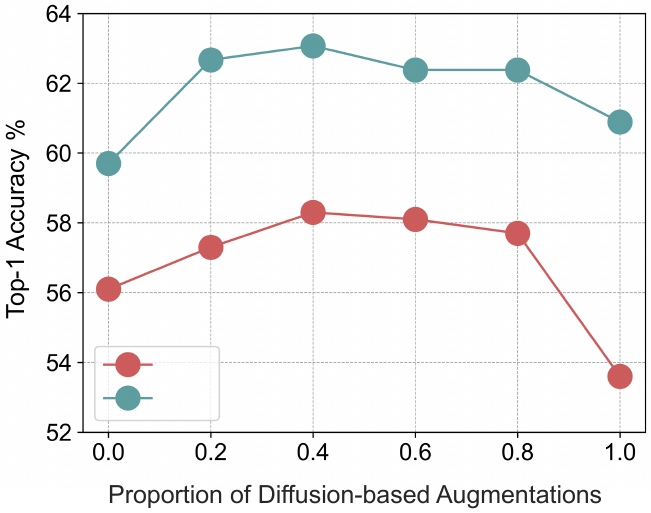}
         \put(-142,40){ \small$\mathcal{S}_1$}
         \put(-142,30){ \small$\mathcal{S}_2$}
	\end{center}
    \vspace{-8pt}
	\captionsetup{font=small}
        \caption{\textbf{Top 1 accuracy} versus the varied proportion of the \textit{standard} augmented views and the \textit{diffusion-based} augmented images  under (a) $\mathcal{S}_1$ and (b) $\mathcal{S}_2$.}
	% \vspace{-10pt}
	\label{fig:A2}
\end{figure*}

\section{Proportion Analysis of the Different Augmented Images}\label{sec:C}

Here, we plot the accuracy with different augmentation ratios of the standard augmentation~\cite{shu2022test} and diffusion-based augmentation~\cite{shu2022test} on the $\mathcal{S}_1$ and $\mathcal{S}_2$ with a backbone of ResNet-50 in Fig.~\ref{fig:A2}. It can be seen from the figure that as the size of the diffusion-based augmentation increases, the performance in the $\mathcal{S}_1$ will also increase, but it will begin to decline when the ratio reaches $0.4$. 
Similarly, the classification performance in the $\mathcal{S}_2$ will first increase and then decline, especially the difference in the $\mathcal{S}_1$ tends to be more prominent. These results support the hypothesis that our approach absorbs the complementary advantages of standard augmentation and diffusion-based augmentations to exploit their respective merits.

\section{Visualization of the Generated and Filtered Image}\label{sec:B}

\begin{figure*}[!ht]
% \vspace{-10pt}
	\begin{center}
		\includegraphics[width=\linewidth]{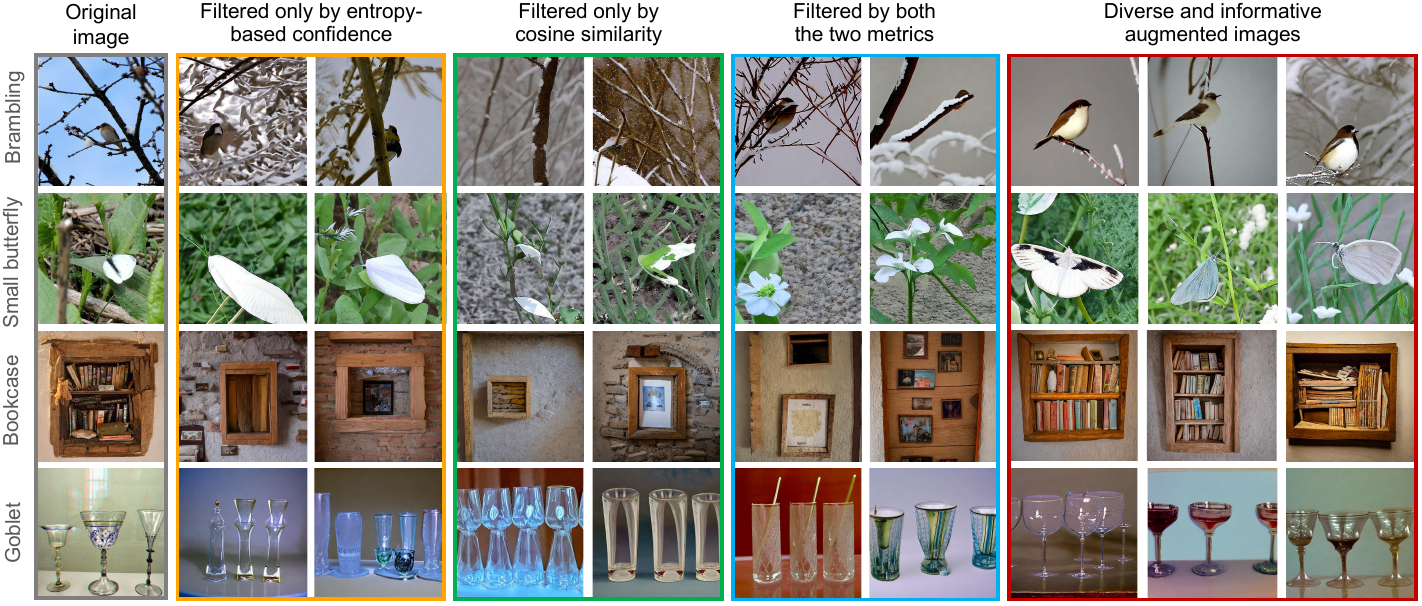}
	\end{center}
	% \vspace{-10pt}
	\captionsetup{font=small}
    	\caption{\textbf{Visualization} of the diverse and informative diffusion-based augmented images and the filtered images, where the \texttt{{\color{gray}gray}} box contains the original test images, the \texttt{{\color{orange}orange}} box contains the images \textit{filtered only by entropy-based confidence}, the \texttt{{\color{darkpastelgreen}green}} box contains the images \textit{filtered only by cosine similarity}, the \texttt{{\color{deepskyblue}blue}} box contains the images \textit{filtered by both the two metrics}, and the \texttt{{\color{firebrick}red}} box contains the diverse and informative diffusion-based augmented images \textit{after filtration}.}
	\label{fig:A1}
\end{figure*}

As we have mentioned, our proposed DiffTPT exploits the advantages of entropy-based confidence selection~\cite{shu2022test} and cosine similarity filtration to improve the adaptability of the model to unknown new test data. Therefore, to display the effectiveness of our method, we visualize the generated images and the filtered images in Fig.~\ref{fig:A1}, where the \texttt{{\color{gray}gray}} box contains the original test images, the \texttt{{\color{orange}orange}} box contains the images filtered only by entropy-based confidence, the \texttt{{\color{darkpastelgreen}green}} box contains the images filtered only by cosine similarity, the \texttt{{\color{deepskyblue}blue}} box contains the images filtered by both the two metrics, and the \texttt{{\color{firebrick}red}} box contains the diverse and informative diffusion-based augmented images after filtration. First, it can be seen from the figure that the diffusion-based augmentation in the red box provides a large number of diverse and informative images while preserving the same key semantics. Second, the images in the green box have deviated from the main target content, indicating that cosine similarity is an effective way to remove spurious augmentation. The other portion of the spurious augmentation in the orange box is deleted by the entropy-based confidence selection mechanism. Therefore, our proposed DiffTPT adopts these two different metrics to take advantage of their complementary merits, thereby being more effective in removing spurious augmentations that are harmful to classification (see the blue box).

% % {\small
% % \bibliographystyle{ieee_fullname}
% % \bibliography{egbib}
% % }

\end{document}